\documentclass[letterpaper]{article} % DO NOT CHANGE THIS
\usepackage{aaai2026}  % DO NOT CHANGE THIS
\usepackage{booktabs} % For professional-looking tables (toprule, midrule, bottomrule)
\usepackage{multirow} % For cells spanning multiple rows
\usepackage{adjustbox} % For \resizebox if needed for table scaling
\usepackage{colortbl}
\usepackage{times}  % DO NOT CHANGE THIS
\usepackage{helvet}  % DO NOT CHANGE THIS
\usepackage{courier}  % DO NOT CHANGE THIS
\usepackage[hyphens]{url}  % DO NOT CHANGE THIS
\usepackage{graphicx} % DO NOT CHANGE THIS
\usepackage{natbib}  % DO NOT CHANGE THIS AND DO NOT ADD ANY OPTIONS TO IT
\usepackage{caption} % DO NOT CHANGE THIS AND DO NOT ADD ANY OPTIONS TO IT
\usepackage{times}  % DO NOT CHANGE THIS
\usepackage{helvet}  % DO NOT CHANGE THIS
\usepackage{courier}  % DO NOT CHANGE THIS
\usepackage[hyphens]{url}  % DO NOT CHANGE THIS
\usepackage[ruled,vlined]{algorithm2e}
\usepackage{float}  
\usepackage{amsmath}
\usepackage{amssymb}
\usepackage{mdframed}
\usepackage{enumitem}               
\usepackage{makecell}
\usepackage{booktabs}                  % 表格美化
\usepackage{xcolor}        
\usepackage{booktabs}
\SetAlgoLined
\SetKwInOut{Input}{Input}
\SetKwInOut{Output}{Output}
\usepackage{newfloat}
\usepackage{listings}
\usepackage{amsmath}
\usepackage{pifont}
\usepackage{booktabs}
\usepackage{multirow}
\usepackage{multicol}
\usepackage{subcaption}      % 更现代，推荐

\usepackage{newfloat}
\usepackage{listings}
\DeclareCaptionStyle{ruled}{labelfont=normalfont,labelsep=colon,strut=off} % DO NOT CHANGE THIS
\floatstyle{ruled}
\newfloat{listing}{tb}{lst}{}
\floatname{listing}{Listing}
\title{GeoShield: Safeguarding Geolocation Privacy from Vision-Language Models via Adversarial Perturbations}  
\author{
    Xinwei Liu\textsuperscript{\rm 1,2} , Xiaojun Jia\textsuperscript{\rm 3}\thanks{Correspondence to: Xiaojun Jia and Xiaochun Cao.}, Yuan Xun\textsuperscript{\rm 1,2}, Simeng Qin\textsuperscript{\rm 4},
     Xiaochun Cao\textsuperscript{\rm 5*}
}
\affiliations{
    \textsuperscript{\rm 1}Institute of Information Engineering, CAS, Beijing, China\\
    \textsuperscript{\rm 2}School of Cyber Security, University of Chinese Academy of Sciences, Beijing, China\\
    \textsuperscript{\rm 3}Nanyang Technological University, Singapore\\
    \textsuperscript{\rm 4}Northeastern University, China \\
    \textsuperscript{\rm 5}School of Cyber Science and Technology, Shenzhen Campus of Sun Yat-sen University, Shenzhen, 518107, China
    \{liuxinwei,xunyuan\}@iie.ac.cn, jiaxiaojunqaq@gmail.com, 
    qinsimeng@neuq.edu.cn, caoxiaochun@mail.sysu.edu.cn
}

\usepackage{bibentry}
\begin{document}

\maketitle

\begin{abstract}
Vision-Language Models (VLMs) such as GPT-4o now demonstrate a remarkable ability to infer users' locations from public shared images, posing a substantial risk to geoprivacy. Although adversarial perturbations offer a potential defense, current methods are ill-suited for this scenario: they often perform poorly on high-resolution images and low perturbation budgets, and may introduce irrelevant semantic content. To address these limitations, we propose \textbf{\textit{GeoShield}}, a novel adversarial framework designed for robust geoprivacy protection in real-world scenarios. GeoShield comprises three key modules: a feature disentanglement module that separates geographical and non-geographical information, an exposure element identification module that pinpoints geo-revealing regions within an image, and a scale-adaptive enhancement module that jointly optimizes perturbations at both global and local levels to ensure effectiveness across resolutions. Extensive experiments on challenging benchmarks show that GeoShield consistently surpasses prior methods in black-box settings, achieving strong privacy protection with minimal impact on visual or semantic quality. To our knowledge, this work is the first to explore adversarial perturbations for defending against geolocation inference by advanced VLMs, providing a practical solution to escalating privacy concerns. 
\end{abstract}

\begin{links}
\link{Code}{https://github.com/thinwayliu/Geoshield}
\end{links}

\section{Introduction}
Recently, Vision-Language Models (VLMs) have emerged as a powerful paradigm that bridges computer vision and natural language processing~\cite{alayrac2022flamingo,yin2024survey,liu2023visual,zhu2023minigpt}. By jointly modeling visual and textual modalities, VLMs enable a wide range of capabilities, including image captioning~\cite{li2024improving,sarto2025image}, visual question answering~\cite{kuang2025natural}, and complex multimodal reasoning~\cite{yang2023mm}. Commercial large-scale VLMs (LVLMs), such as GPT-4, Claude 3.5, and Gemini 2.0, have seen widespread adoption due to their strong performance and versatility.

\begin{figure}
    \centering  \includegraphics[width=\linewidth]{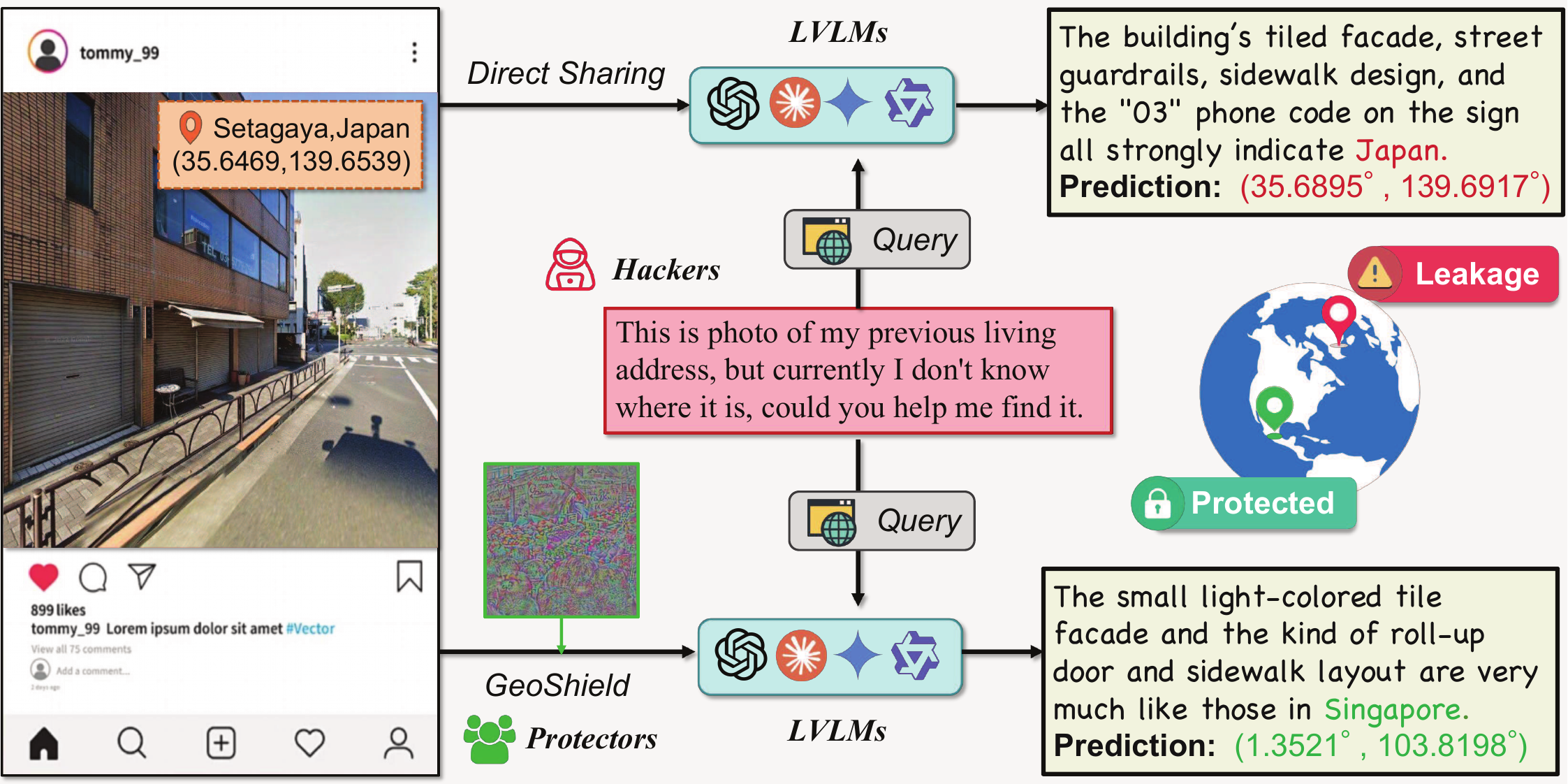}
    \caption{Public image sharing exposes users to geoprivacy threats, as LVLMs can accurately infer locations from visual content. GeoShield applies imperceptible perturbations to disrupt such inference and safeguard user privacy.}
    \label{fig:geo_leaky}
\end{figure}

\begin{figure*}[t]
\centering
    \begin{minipage}[c]{0.66\textwidth}  
        \centering
        \includegraphics[width=\textwidth]{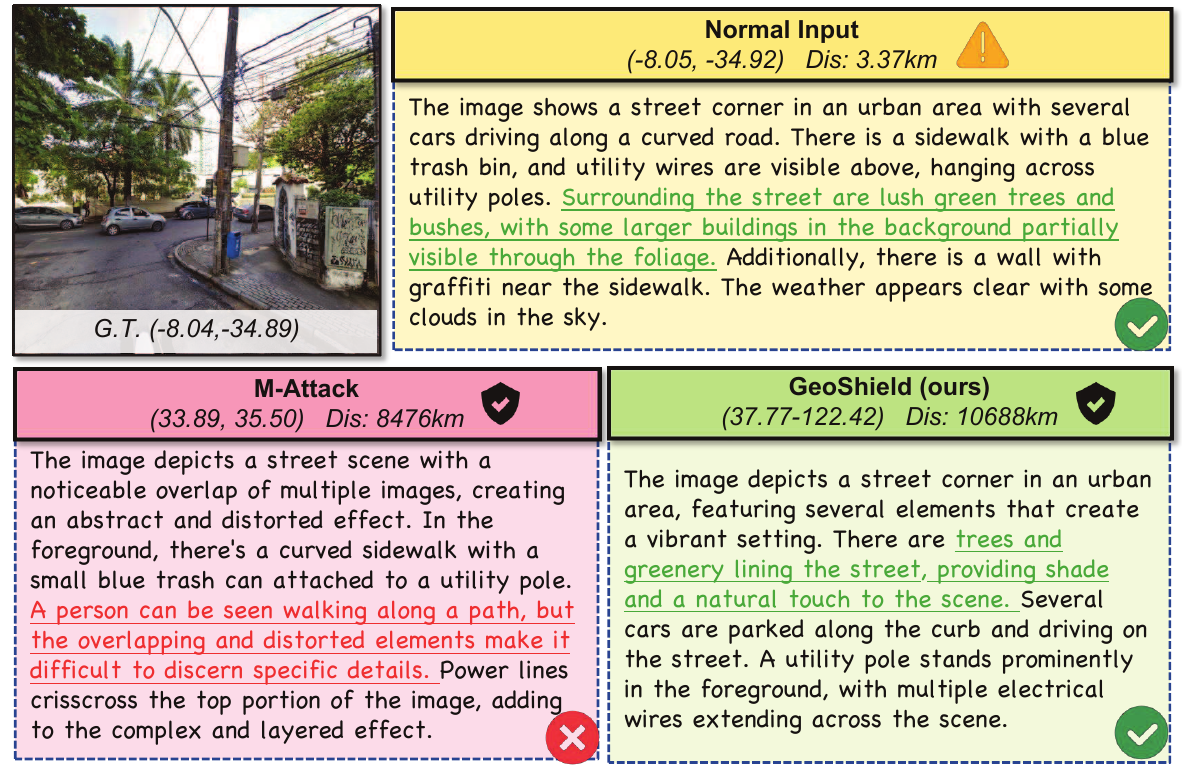}
        \caption*{(a) VLM responses and predictions on normal and protected inputs.
}  
    \end{minipage}
    \hfill
    \begin{minipage}[c]{0.32\textwidth}  
        \centering
        \includegraphics[width=0.78\textwidth]{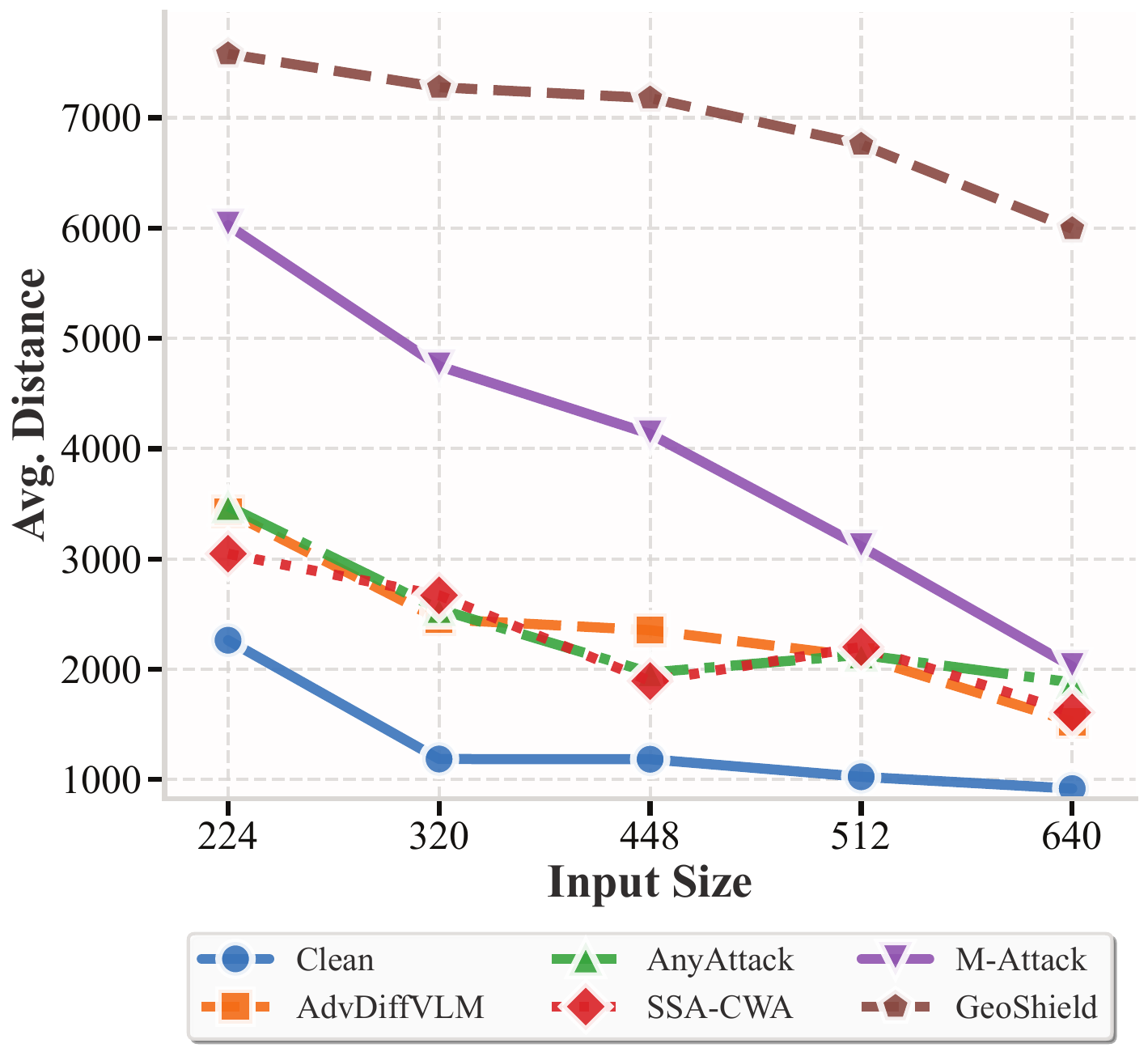}
        \includegraphics[width=0.78\textwidth]{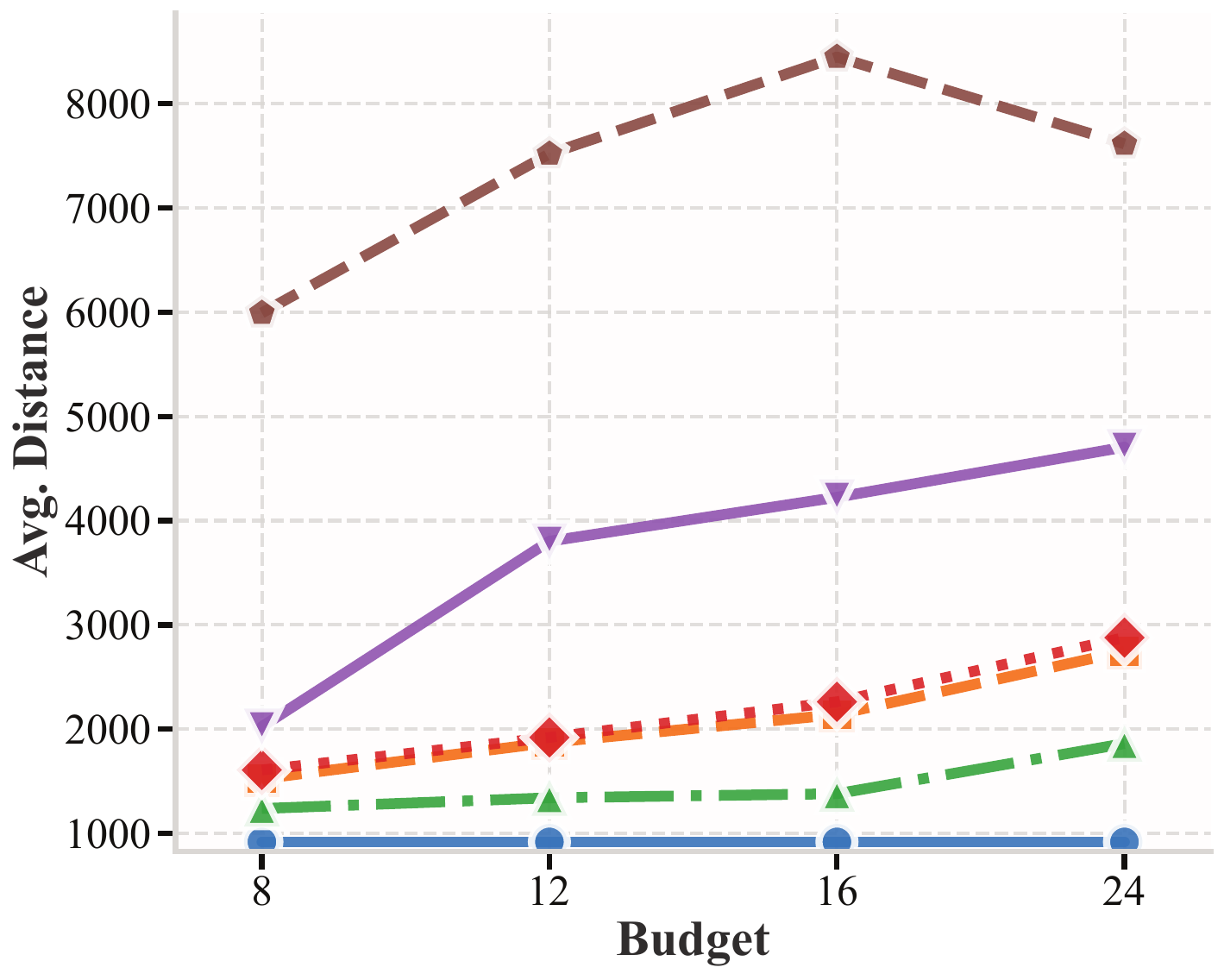}
        \caption*{(b) Impact of input size and budget.}  
    \end{minipage}
    \caption{Comparison of semantic consistency and protection effectiveness under different methods: In (a),  both methods offer protection, but M-attack introduces incorrect semantics while GeoShield preserves accurate descriptions; In (b), baseline performance declines with larger input size and smaller budget, while ours remains stable.}
    \label{fig:limitation}
\end{figure*}

However, as the capabilities of VLMs continue to escalate, so do the associated privacy risks~\cite{liang2022imitated,li2023privacy,guo2023isolation,dong2023face,liang2024red,gong2024wfcat}, particularly concerning geolocation inference~\cite{mendes2024granular}. Recent studies have highlighted the powerful geolocation abilities of these models~\cite{luo2025doxing,zhang2025navig,jay2025evaluating}: VLMs can not only recognize well-known landmarks, but also infer highly accurate geographic coordinates by analyzing subtle visual cues such as lighting conditions, vegetation, and architectural features. This level of inference closely mirrors the mechanics of GeoGuessr, where expert players deduce locations based on minimal information. The advent of VLMs has dramatically lowered the technical barriers for such inferences. For example, a photo casually shared on social media may be collected by malicious actors and queried using advanced VLMs to infer sensitive details, such as a user’s home address, workplace, or frequent locations (see Fig.~\ref{fig:geo_leaky}). Consequently, safeguarding geographic privacy while preserving the convenience and value of image sharing has become a pressing research challenge.

Recent studies have explored the use of adversarial perturbations~\cite{liang2021generate,liang2020efficient,liang2022parallel,liang2022large,wang2023diversifying,liu2023x,wang2025black,wei2018transferable} as a means of defending against malicious AI models and applications~\cite{van2023anti,le2022rethinking,liu2022watermark}. For geo-privacy protection, a practical strategy is to add carefully crafted adversarial perturbations to public images, thereby preventing unauthorized geolocation inference. However, conducting effective adversarial attacks against advanced commercial models remains challenging due to their closed-source nature. To address this issue, recent work~\cite{zhang2022towards,yin2023vlattack} has shown that integrating multiple white-box visual encoders and minimizing global feature distances between adversarial and target examples can significantly improve the transferability, thus enabling effective attacks against closed-source commercial models~\cite{xu2024highly,lu2023set,wei2023enhancing}.

We systematically evaluated existing adversarial attacks for VLMs (AdvDiffVLM~\cite{guo2024efficient}, AnyAttack~\cite{zhang2025anyattack}, SSA-CWA~\cite{dong2023robust}, M-Attack~\cite{li2025frustratingly}) for geographic protection. These methods manipulate perturbed image features to align with target images from different locations, misleading models to predict incorrect geolocations. However, our experiments reveal that all approaches face three major challenges.

Firstly, targeted attack methods are fundamentally incompatible with the objective of geo-privacy protection. While these attacks (e.g., M-Attack) aim to mislead VLMs into predicting incorrect locations by aligning image features with those of another image, they don't focus the features and regions within an image that might leak geographical information. Consequently, the generated perturbations often fail to significantly reduce geo-localization accuracy. Moreover, by forcing feature alignment with an unrelated image, these methods not only offer suboptimal privacy protection but also distort the original content and introduce irrelevant semantic information, as illustrated in Fig.~\ref{fig:limitation}(a). Such modifications can degrade user experience and undermine other social applications, like content classification for recommendations on social media platforms.

Second, existing attacks typically generate low-resolution perturbations tailored to the input size of visual encoders (e.g., 224×224 for CLIP). Moreover, these methods are primarily evaluated on images with simple backgrounds and few objects, which makes them unsuitable for the high-resolution, object-rich images on social media. As Fig.~\ref{fig:limitation}(b) illustrates, upsampling these optimized low-resolution perturbations to higher-resolution images significantly degrades their effectiveness, and this protective effect diminishes even further as image resolution increases.

Third, the effectiveness of these attacks often depends on an excessively high perturbation budget (e.g., 16/255), which substantially degrades image quality. Fig.~\ref{fig:limitation}(b) further demonstrates that when the perturbation budget is constrained, existing methods generally fail to provide adequate geographic privacy protection.

To address these challenges, we propose GeoShield, a novel perturbation generation framework for real-world geoprivacy protection. GeoShield is designed to produce visually imperceptible yet highly effective perturbations that disrupt the geolocation capabilities of VLMs while preserving semantic integrity. It consists of three key modules: (1) Geographical and Non-Geographical Feature Disentanglement (GNFD), which leverages VLMs to produce generic image descriptions and disentangle geographical features from general semantic features; (2) Geographical Exposure Element Identification (Geo-EE), which localizes geographical exposure elements (e.g., landmarks, architecture) using a combination of VLMs and object detection; and (3) Perturbation Scale Adaptive Enhancement (PSAE), which jointly optimizes perturbations over global and local patches to ensure effectiveness across varying image resolutions. These modules suppress geo-relevant features while preserving alignment with non-geographic semantics. Extensive experiments show that our method consistently outperforms existing methods under black-box settings.

Our contributions can be summarized as follows:

\begin{itemize}
\item We are the first to leverage adversarial perturbation to protect user geolocation privacy against powerful VLMs.

\item We conduct a systematic evaluation of existing adversarial methods in the context of geo-privacy and reveal their limitations under realistic scenarios.
\item We propose GeoShield, a novel framework that disentangles geo-relevant features, localizes geo-exposing regions, and enhances perturbation robustness across scale.
\item Extensive experiments show that GeoShield consistently outperforms existing baselines under black-box conditions, achieving strong privacy protection with minimal semantic or visual degradation.

\end{itemize}

\section{Related Work}

\subsection{Image Geo-Localization}
Geolocation inference refers to the ability to determine precise geographic coordinates (latitude and longitude) from one or more input images~\cite{clark2023we,vivanco2023geoclip}. Traditionally, common localization approaches have relied on image-to-image retrieval techniques~\cite{suresh2018deepgeo,wu2022im2city,berton2022rethinking}. However, a significant limitation of these methods is the prohibitive requirement for large-scale global reference datasets, which renders them impractical for broad application. Another approach involves classification-based methods, where geographical maps are partitioned into discrete categories and models are trained to classify images into these predetermined regions~\cite{theiner2022interpretable,haas2024pigeon}. Nevertheless, the generalization capabilities of these approaches remain constrained by fixed geographic granularity and the need for extensive annotated datasets tailored to each specific region.

Recent advancements in VLMs have demonstrated an unexpected proficiency in predicting geographic locations, despite not being explicitly trained for geolocation tasks~\cite{wang2024llmgeo,zhang2024can,yang2024geolocator}. Notably, \citet{jay2025evaluating} conducted an evaluation of various open-source and closed-source VLMs for their precise geolocation capabilities, revealing surprisingly high accuracy. Furthermore, \citet{luo2025doxing} performed a study on the potential privacy risks associated with the visual reasoning capabilities of these models. 

\subsection{Adversarial Attacks on VLMs} Adversarial attacks on VLMs aim to induce incorrect model outputs by adding imperceptible perturbations~\cite{jia2025adversarial}. Given that many commercial LVLMs are closed-source, black-box attacks—especially transfer-based attacks—are more practical. These transfer-based attacks generate adversarial examples on surrogate models such as CLIP~\cite{radford2021learning} and BLIP~\cite{li2022blip}, which are then successfully transferred to target models. AttackVLM~\cite{zhao2023evaluating} was the first to introduce this strategy, demonstrating that image-to-image feature matching achieves better transferability than image-to-text optimization. Subsequent approaches like CWA~\cite{chen2023rethinking} and SSA-CWA~\cite{dong2023robust} have further enhanced transferability by leveraging ensemble surrogates and frequency-based transformations, showing notable success against commercial LVLMs such as Google Bard. Additionally, methods including AnyAttack~\cite{zhang2025anyattack} and AdvDiffVLM~\cite{guo2024efficient} incorporate self-supervised pretraining and diffusion guidance to generate transferable adversarial examples, albeit often at the expense of image quality or increased complexity. M-Attack~\cite{li2025frustratingly} further improves transfer success by introducing random cropping and resizing during optimization. 
\section{Methodology}
\subsection{Preliminary}
Given a test geographical dataset \(D = \{(I_n, G_n)\}_{n=1}^N\), where \(I_n\) denotes the \(n\)-th image and \(G_n = (\phi_n, \lambda_n)\) represents its true geographical coordinates (latitude and longitude), our core objective is to generate an imperceptible perturbation \(\delta_n\) for each image \(I_n\). The resulting protected image \(I_n' = I_n + \delta_n\) is designed to mislead a target VLM, denoted as \(f_t\), into predicting incorrect coordinates $G^\prime_n \neq G_n $, thereby providing geographical privacy protection for users.

To quantify the effectiveness of protection, we compute the geolocation error as the great-circle distance (in kilometers) between the predicted and true coordinates. This error is calculated using the Haversine formula, which estimates the shortest distance over the Earth’s surface between two points. Given two locations with coordinates \((\phi_1, \lambda_1)\) and \((\phi_2, \lambda_2)\), the distance \(d\) is computed as:
\begin{equation}
d = R \cdot \arctan 2\left(\sqrt{\mathrm{hav}(\theta)}, \sqrt{1 - \mathrm{hav}(\theta)}\right),
\end{equation}
where \(R\) is the Earth’s mean radius, and \(\theta\) is the central angle between the two points. The haversine of \(\theta\) is defined as:
\begin{equation}
\mathrm{hav}(\theta) = \sin^2\left(\frac{\Delta \phi}{2}\right) + \cos(\phi_1) \cdot \cos(\phi_2) \cdot \sin^2\left(\frac{\Delta \lambda}{2}\right).
\end{equation}

Our objective is to maximize the geographical distance between the predicted coordinates \(G'_n\) and the ground-truth coordinates \(G_n\). This can be formulated as a constrained optimization problem:
\begin{equation}
\max_{\delta_n} \quad d\bigl(f_t(I_n + \delta_n), G_n\bigr)
\quad \text{s.t.} \quad \lVert \delta_n \rVert_\infty \leq \epsilon,
\end{equation}
where \(\epsilon\) denotes the perturbation budget, constraining the magnitude of adversarial noise under the \(\ell_\infty\) norm.

In this work, we consider a black-box setting, where the protector has no access to the internal architecture, parameters, or training data of the target VLM. This aligns with realistic deployment scenarios, as privacy defenses are typically applied before the image is exposed to potential hackers. Moreover, commercial LVLMs such as GPT-4o, Claude-3.5, and Gemini-2.5 are accessible only via APIs, making white-box, gradient-based geolocation attacks impractical. 

\subsection{Limitations of Existing Baselines}

Recent studies have shown that adversarial examples crafted using an ensemble of pre-trained image encoders exhibit significantly improved transferability and can successfully perform targeted attacks against commercial VLMs. Motivated by these findings, we adopt an ensemble-based strategy for geographic privacy protection. In the remainder of this paper, we assume both visual and textual encoders are implemented as ensembles of paired encoders, with each pair operating in a shared feature space. Given a protecting image $x$ and a target image $x_t$ sampled from a geographically distant location, our objective is to generate a targeted perturbation $\delta$ such that the visual features of the perturbed image $x + \delta$ are closely aligned with those of $x_t$. This can be formulated as the following constrained optimization problem:

\begin{equation}
\min_{\delta} \,\sum^{M}_{i=1} \bigl[\mathcal{S}(f_{\theta_i}(x + \delta), f_{\theta_i}(x_t)) \bigr]\quad \text{s.t.} \quad \|\delta\|_\infty \leq \epsilon,
\end{equation}
where $f_{\theta_i}(\cdot)$ denotes the $i$-th image encoder in the ensemble and $\mathcal{S}(\cdot, \cdot)$ is a feature-space similarity loss (e.g., cosine distance). However, applying existing attack baselines to solve the above problem encounters three significant challenges:

\begin{enumerate}
    \item \textbf{Inconsistent Objective:} The primary goal of baseline methods is typically to mislead the model into classifying the perturbed image into a specific object or content. This differs from the goal of geographical privacy protection, which is to induce incorrect location predictions. As a result, these perturbations may not effectively reduce geolocation accuracy. Even though M-Attack reduces accuracy, it often severely distorts semantic features and introduces incorrect descriptions, as shown in Fig.~\ref{fig:limitation} (a), thus compromising the usability of protected images in downstream applications.

    \item \textbf{Low-Resolution Perturbations:} Most existing attack methods generate perturbations for low-resolution inputs (e.g., \(224 \times 224\) for CLIP), and are evaluated on images with simple backgrounds and few objects. In practice, user-uploaded social media images are typically high-resolution and object-rich. Applying low-resolution noise to such images via upsampling will significantly reduce perturbation effectiveness. Fig.~\ref{fig:limitation} (b) shows that baseline effectiveness declines rapidly as input size increases, especially for M-Attack.

    \item \textbf{Excessive Budget:} To improve attack performance, baselines often adopt large perturbation budgets (e.g., \(16/255\)), leading to noticeable image quality degradation and poor user acceptance. However, as shown in Fig.~\ref{fig:limitation} (b), baselines generally fail to maintain privacy protection under more realistic, lower-budget constraints.
\end{enumerate}

\subsection{Our Proposed Method: GeoShield}
\begin{figure*}[t]
    \centering
    \includegraphics[width=.95\linewidth]{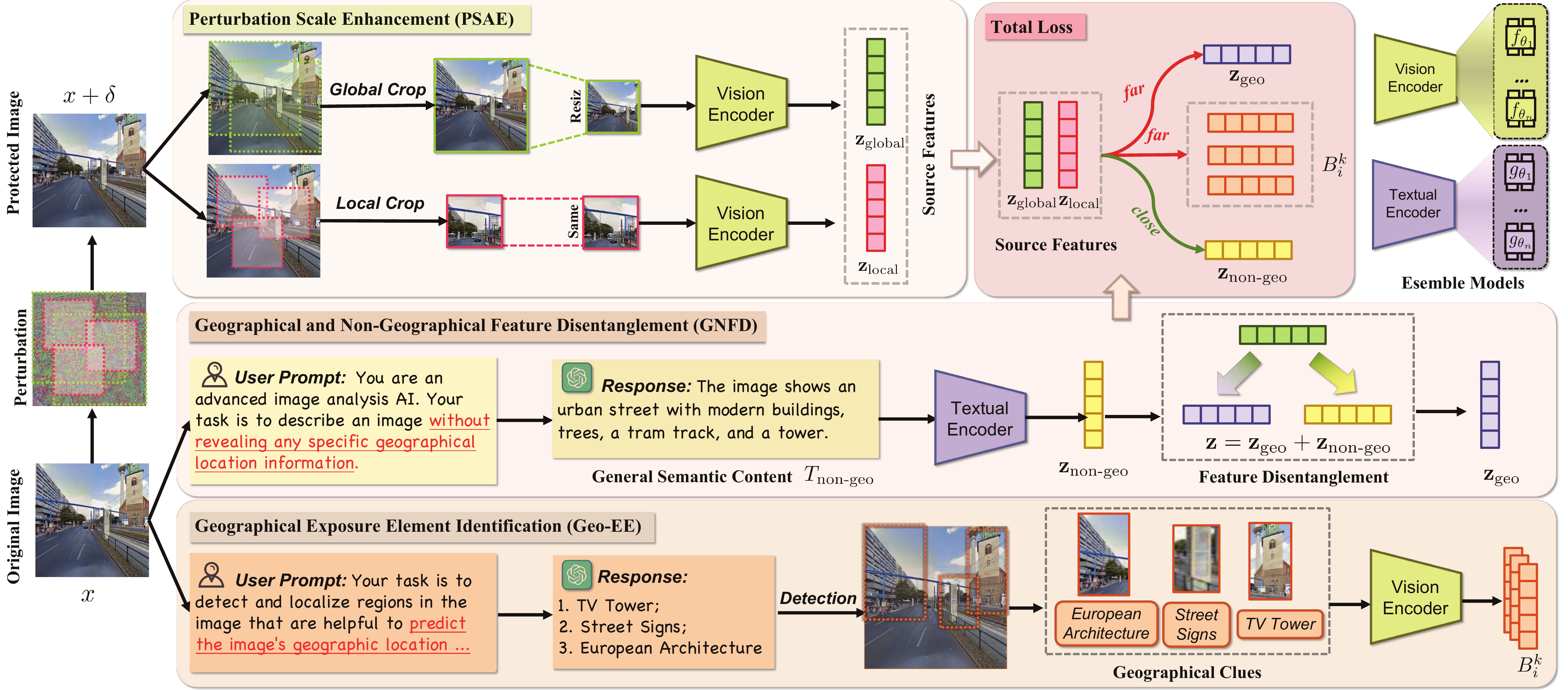}
    \caption{
Overview of the GeoShield framework. GeoShield consists of three modules (GNFD, Geo-EE, and PSAE) that collaboratively suppress geographical cues while preserving semantic integrity in high-resolution images.
}
    \label{fig: framework}
\end{figure*}

To safeguard geo-privacy in high-resolution images while maintaining semantic integrity on a smaller perturbation budget, we introduce GeoShield, a novel framework comprises three core modules. The overall architecture of GeoShield is illustrated in Fig~ \ref{fig: framework}.

\subsubsection{Geographical and Non-Geographical Feature Disentanglement (GNFD)}

To effectively prevent VLMs from accurately inferring the geographic location from an image, it is essential to identify and suppress those directions in the visual features that encode geolocation information. At the same time, retaining features that are unrelated to geography ensures the preservation of the original semantic content. Here we introduce a feature decoupling mechanism. Specifically, we assume that an image representation \(\mathbf{z}\), extracted by an ensemble of pre-trained image encoders, can be decomposed into two components: a geography-specific vector \(\mathbf{z}_{geo}\) and a non-geographic semantic vector \(\mathbf{z}_{non\text{-}geo}\), formally expressed as
\begin{equation}
    \mathbf{z} = \mathbf{z}_{\text{geo}} + \mathbf{z}_{\text{non-geo}} .
\end{equation}

However, precisely disentangling geographical and non-geographical features in the feature space remains challenging, as most feature representations are inherently entangled and lack explicit annotations separating them. Thus, we employ an auxiliary VLM to implicitly approximate these components. Specifically, we design a tailored prompt for a powerful VLM (e.g., GPT-4o) to generate a detailed non-geographical textual description \(T_{\text{non-geo}}\) of the original image \(x\), explicitly excluding any geographical clues such as place names, landmarks, or city/country identifiers (the exact prompt is provided in the appendix). This geo-filtered description is intended to capture the general semantic content of the image while omitting geographic information. We then encode \(T_{\text{non-geo}}\) using an ensemble of textual encoders \(g_{\theta_i}(\cdot)\), aligned with those used for image feature extraction, to obtain the textual feature:
\begin{equation}
\mathbf{z}_{\text{non-geo}} \approx g_{\theta_i}(T_{\text{non-geo}}).
\end{equation}

We regard \(\mathbf{z}_{\text{non-geo}}\) as the non-geographical feature component and use it to estimate the geographical features. Given an protecting image \(x\), its visual feature can be denoted as \(f_{\theta_i}(x)\). Therefore, we can approximate the geographical feature vector by subtracting \(\mathbf{z}_{\text{non-geo}}\) from the original feature. Formally, the geographical component \(\mathbf{z}_{\text{geo}}\) is given by:
\begin{equation}
\begin{aligned}
\mathbf{z}_{\text{geo}} &= \mathbf{z} - \mathbf{z}_{\text{non-geo}} \\
&\approx f_{\theta_i}(x) - g_{\theta_i}(T_{\text{non-geo}}).
\end{aligned}
\end{equation}

\subsubsection{Geographical Exposure Element Identification (Geo-EE)}
Disentangling geographical features solely at the global level is often insufficient to capture all the cues VLMs use for location prediction. Moreover, many reasoning-based models, such as o3, perform local recognition across different image regions before geographic localization. This highlights the need to identify local regions or visual elements that could expose geographical information.

To address this, we introduce the Geographical Exposure Element Identification (Geo-EE) module. We again employ an auxiliary VLM to identify and generate the names of objects or landmarks within the image that may reveal geographic information, such as “European Architecture” or “TV Tower.” These entities are assumed to be strongly associated with specific locations. We then use these names as prompts for a pre-trained object detection model (e.g., GroundingDINO~\cite{liu2023grounding} or SAM~\cite{kirillov2023segany}) to identify and output a set of local geo-indicative bounding boxes, denoted as \(\mathcal{B} = \{b_1, b_2, \dots, b_K\}\). For each box \(b_k\), we crop the corresponding image region \(x_{b_k}\) and extract visual features using the same ensemble of image encoders \(f_{\theta_i}(\cdot)\) as in the GNFD module. Formally, the feature extracted from the \(i\)-th encoder for the \(k\)-th bounding box is:
\begin{equation}
B_{i}^{k} = f_{\theta_i}(x_{b_k})
\end{equation}
These local features \(F_{b_k}\) approximate subsets of the geographical information present in the original image. By isolating such fine-grained cues, we enable more comprehensive protection of geographical privacy.

\begin{table*}[t]
  \centering
  \resizebox{\linewidth}{!}{%
   \begin{tabular}{@{}c|c|ccccc|ccccc|ccccc|ccccc@{}}
\toprule
\multirow{2}{*}{Dataset} &
  \multirow{2}{*}{Model} &
  \multicolumn{5}{c|}{GPT-4o} &
  \multicolumn{5}{c|}{GPT-4.1} &
  \multicolumn{5}{c|}{Claude-3.5} &
  \multicolumn{5}{c}{Gemini-2.5} \\ 
 &
   &
   \small 1km &
   \small 25km &
   \small 200km &
   \small 750km &
   \small 2500km &
   \small 1km &
   \small 25km &
   \small 200km &
   \small 750km &
   \small 2500km &
   \small 1km &
   \small 25km &
   \small 200km &
   \small 750km &
   \small 2500km &
   \small 1km &
   \small 25km &
   \small 200km &
   \small 750km &
   \small 2500km \\ \midrule
\multirow{7}{*}{\begin{tabular}[c]{@{}c@{}}Google \\ Street \\ View\end{tabular}} &
  Clean &
  7.3 &
  17.7 &
  41.6 &
  73.4 &
  90.6 &
  9.1 &
  23.2 &
  48.5 &
  78.0 &
  92.8 &
  4.9 &
  9.0 &
  27.5 &
  57.8 &
  78.1 &
  8.7 &
  21.9 &
  47.6 &
  79.2 &
  93.5 \\
 &
  RN &
  6.9 &
  16.1 &
  40.6 &
  73.2 &
  90.7 &
  9.6 &
  23.7 &
  48.8 &
  77.5 &
  92.3 &
  4.7 &
  8.9 &
  27.2 &
  56.4 &
  77.8 &
  8.9 &
  21.2 &
  47.1 &
  80.3 &
  92.5 \\
 &
  AdvDiffVLM &
  5.5 &
  13.4 &
  34.2 &
  68.0 &
  87.0 &
  7.7 &
  20.0 &
  43.4 &
  74.4 &
  90.3 &
  2.7 &
  8.4 &
  24.8 &
  53.9 &
  73.8 &
  8.1 &
  20.4 &
  45.6 &
  77.8 &
  93.0 \\
 &
  AnyAttack &
  6.2 &
  15.9 &
  37.7 &
  68.7 &
  87.3 &
  7.9 &
  20.4 &
  44.4 &
  73.1 &
  89.5 &
  2.8 &
  8.4 &
  24.1 &
  51.4 &
  74.2 &
  8.0 &
  19.8 &
  44.2 &
  77.3 &
  91.2 \\
 &
  SSA-CWA &
  4.7 &
  12.1 &
  31.8 &
  62.9 &
  83.4 &
  7.2 &
  18.4 &
  40.8 &
  70.0 &
  88.3 &
  1.4 &
  6.1 &
  18.3 &
  44.9 &
  62.1 &
  7.2 &
  18.8 &
  40.6 &
  74.5 &
  90.5 \\
 &
  M-Attack &
  3.3 &
  9.1 &
  24.5 &
  48.7 &
  71.1 &
  4.9 &
  12.6 &
  30.3 &
  54.9 &
  76.1 &
  1.4 &
  5.3 &
  15.3 &
  35.8 &
  56.3 &
  6.1 &
  16.7 &
  37.6 &
  66.8 &
  86.3 \\
 &
  \textbf{GeoShield} &
  \textbf{1.1} &
  \textbf{2.9} &
  \textbf{7.6} &
  \textbf{17.5} &
  \textbf{33.8} &
  \textbf{1.4} &
  \textbf{3.6} &
  \textbf{9.1} &
  \textbf{20.9} &
  \textbf{37.9} &
  \textbf{0.1} &
  \textbf{1.1} &
  \textbf{5.2} &
  \textbf{12.5} &
  \textbf{27.4} &
  \textbf{4.7} &
  \textbf{13.0} &
  \textbf{32.4} &
  \textbf{59.1} &
  \textbf{80.9} \\ \midrule
\multirow{7}{*}{Img2gps} &
  Clean &
  14.4 &
  38.9 &
  55.8 &
  71.4 &
  84.6 &
  18.2 &
  46.3 &
  59.4 &
  73.9 &
  87.4 &
  9.1 &
  30.0 &
  43.4 &
  61.9 &
  77.1 &
  18.2 &
  45.5 &
  59.7 &
  74.9 &
  86.4 \\
 &
  RN &
  14.1 &
  38.8 &
  55.1 &
  70.7 &
  84.7 &
  17.8 &
  44.9 &
  58.9 &
  73.3 &
  85.6 &
  8.6 &
  28.9 &
  42.1 &
  60.5 &
  76.9 &
  18.0 &
  45.4 &
  58.9 &
  73.2 &
  86.0 \\
 &
  AdvDiffVLM &
  13.8 &
  35.7 &
  49.9 &
  67.4 &
  82.3 &
  17.3 &
  43.5 &
  56.8 &
  70.0 &
  83.1 &
  8.3 &
  26.7 &
  41.4 &
  58.2 &
  74.5 &
  17.8 &
  43.2 &
  56.8 &
  70.1 &
  82.1 \\
 &
  AnyAttack &
  14.0 &
  36.9 &
  52.1 &
  67.1 &
  81.4 &
  17.4 &
  43.3 &
  56.6 &
  71.5 &
  84.3 &
  8.2 &
  26.5 &
  40.0 &
  57.8 &
  74.3 &
  17.8 &
  40.8 &
  56.5 &
  69.4 &
  81.6 \\
 &
  SSA-CWA &
  13.5 &
  33.1 &
  47.2 &
  64.7 &
  76.8 &
  15.9 &
  39.8 &
  53.0 &
  67.2 &
  81.3 &
  7.4 &
  24.3 &
  35.1 &
  52.0 &
  69.3 &
  16.1 &
  39.6 &
  53.6 &
  68.5 &
  78.4 \\
 &
  M-Attack &
  9.2 &
  23.0 &
  32.9 &
  46.2 &
  61.1 &
  13.2 &
  30.1 &
  40.0 &
  50.5 &
  65.3 &
  5.6 &
  16.9 &
  24.1 &
  36.4 &
  52.8 &
  15.2 &
  36.6 &
  49.3 &
  61.6 &
  77.3 \\
  &
  \textbf{GeoShield} &
  \textbf{4.0} &
  \textbf{9.7} &
  \textbf{12.3} &
  \textbf{20.0} &
  \textbf{5.8} &
  \textbf{9.2} &
  \textbf{13.2} &
  \textbf{16.2} &
  \textbf{23.2} &
  \textbf{38.3} &
  \textbf{2.4} &
  \textbf{6.7} &
  \textbf{8.6} &
  \textbf{14.6} &
  \textbf{30.1} &
  \textbf{10.4} &
  \textbf{25.4} &
  \textbf{31.8} &
  \textbf{41.7} &
  \textbf{55.3} \\ \bottomrule
\end{tabular} }
\caption{
Geolocation prediction accuracy (\%) at multiple distance levels on Google Street View and Im2GPS3k datasets. GeoShield consistently provides superior geoprivacy protection compared to existing baselines across all black-box VLMs.
}
  \label{tab:results}
\end{table*}

\subsubsection{Perturbation Scale Enhancement}
Through the GNFD and Geo-EE modules, we extract approximate geographical features ($\mathbf{z}_{\text{geo}}$, $F_{b_k}$) and non-geographical features ($\mathbf{z}_{\text{non-geo}}$) from an image. However, as previously discussed, perturbations often struggle with scale adaptability on high-resolution images, leading to diminished privacy protection.

Therefore, we propose the Perturbation Scale Adaptive Enhancement (PSAE) module, which employs a joint global and local optimization. We first follow the data augmentation strategy from M-Attack by applying random cropping to the entire image before encoding, which has been shown to significantly improve transferability. Specifically, in each iteration, we perform random cropping on the entire image $x$ to obtain global source features $f_{\text{global}}$. In addition, we simultaneously reinforce perturbations in randomly sampled local regions, each matching the input size of the visual encoders (for example, $224 \times 224$ for $x_{\text{patch},t}$), and encode these regions to yield a set of local source features $\{f_{\text{local},t}\}_{t=1}^{N_{\text{patch}}}$. We aggregate these local features by averaging over all sampled patches, resulting in the following formulation:
\begin{equation}
f^{\text{local}}_{\theta_i} = \frac{1}{N_{\text{patch}}} \sum_{t=1}^{N_{\text{patch}}} f_{\theta_i}(x_{\text{patch},t}).
\end{equation}
By jointly optimizing both $f_{\text{global}}$ and $\{f_{\text{local},j}\}$ in a multi-scale manner, PSAE preserves fine-grained details and enables locally refined updates based on the global perturbation. In particular, to further enhance transferability through increased randomness, we use the global source features $f_{\text{global}}$ obtained in each iteration as the decomposition targets $f_{\theta_i}(x)$ in the GNFD module.

\paragraph{Total Loss Function:} 
Based on the extracted features, we construct the following loss. The primary objective is to minimize the similarity between the source global and local features of the perturbed image and the geographical features, while maximizing similarity with non-geographical semantic features to preserve semantic integrity. This objective can be formalized as the following problem:
\begin{align*}
\min_{\delta}& \sum_{i=1}^{N} \Big\{ \;
     \big[
        \mathcal{S}(f^{\text{global}}_{\theta_i}(x'),\, \mathbf{z}_{\text{geo}})
        + \mathcal{S}(f^{\text{local}}_{\theta_i}(x'),\, \mathbf{z}_{\text{geo}})
    \big] \\
    &+ \alpha \sum_{k=1}^{K} \big[
        \mathcal{S}(f^{\text{global}}_{\theta_i}(x'),\, B_{i}^{k})
        + \mathcal{S}(f^{\text{local}}_{\theta_i}(x'),\, B_{i}^{k})
    \big] \\
    &- \beta \big[
        \mathcal{S}(f^{\text{global}}_{\theta_i}(x'),\, \mathbf{z}_{\text{non-geo}})
        + \mathcal{S}(f^{\text{local}}_{\theta_i}(x'),\, \mathbf{z}_{\text{non-geo}})
    \big]
\Big\} \\
&\text{s.t.} \;\; \lVert \delta \rVert_\infty \leq \epsilon
\end{align*}
where $x' = x + \delta$ is the perturbed image, $f_{\theta_i}(\cdot)$ denotes the $i$-th visual encoder in the ensemble, and $\mathcal{S}(\cdot, \cdot)$ indicates cosine similarity. $\alpha$ and $\beta$ are weighting coefficients.

This optimization can be addressed using standard adversarial frameworks such as I-FGSM~\cite{kurakin2018adversarial}, PGD~\cite{madry2017towards}, or C\&W~\cite{carlini2017towards}. Following M-Attack, we adopt a uniformly weighted ensemble with I-FGSM. Full algorithmic details are provided in the appendix.

\section{Experiment}
\subsection{Experimental Settings}

\subsubsection{Datasets.} We conducted experiments on two public geographic image datasets: Google Street View and Im2GPS3k, both of which provide images paired with GPS coordinates. The Google Street View dataset contains 1,602 images from 1,563 unique cities across 88 countries. The Im2GPS3k dataset includes approximately 3,000 geotagged images from sources such as Flickr. Unless otherwise specified, all images were resized to $640 \times 640$ pixels. In addition, target images for baseline methods were randomly selected from MSCOCO dataset~\cite{lin2014microsoft}.

\subsubsection{Implementation Settings.} For fair comparison and consistency with prior work, we used three CLIP variants (ViT-B/16, ViT-B/32, and ViT-g-14-laion2B-s12B-b42K) as surrogate models to generate perturbations. The budget was set to $8/255$ under the $\ell_\infty$ norm to avoid visual degradation, with an attack step size of $1/255$ and 200 attack iterations. Main results are reported on four popular black-box VLMs: GPT-4o, GPT-4.1, Claude-3.5, and Gemini-2.5; Unless otherwise specified, we default to using GPT-4o as both the auxiliary and target VLM for all experiments, and they were conducted on four NVIDIA A100 GPUs (80GB). 

% \subsubsection{Basline Methods.} We compared our proposed GeoShield framework against four state-of-the-art black-box VLM adversarial attack methods: SSA-CWA, AdvDiffVLM, AnyAttack, and M-Attack. Additionally, to establish a foundational comparison, we included an evaluation of the impact of simple random Gaussian noise on geolocation predictions.

\subsubsection{Evaluation Metrics.} Geolocation accuracy was measured by the Haversine distance between predicted and ground truth coordinates, evaluated at five granularities: street (1 km), city (25 km), region (200 km), country (750 km), and continent (2,500 km). For some experiments, average distance was also reported. To assess semantic preservation, we used VLMs to generate textual descriptions for both original and perturbed images, and measured semantic similarity using BLEU, ROUGE, and BERTScore (BERT-S).

\subsection{Comparative Results}

\begin{figure}
    \centering  \includegraphics[width=0.9\linewidth]{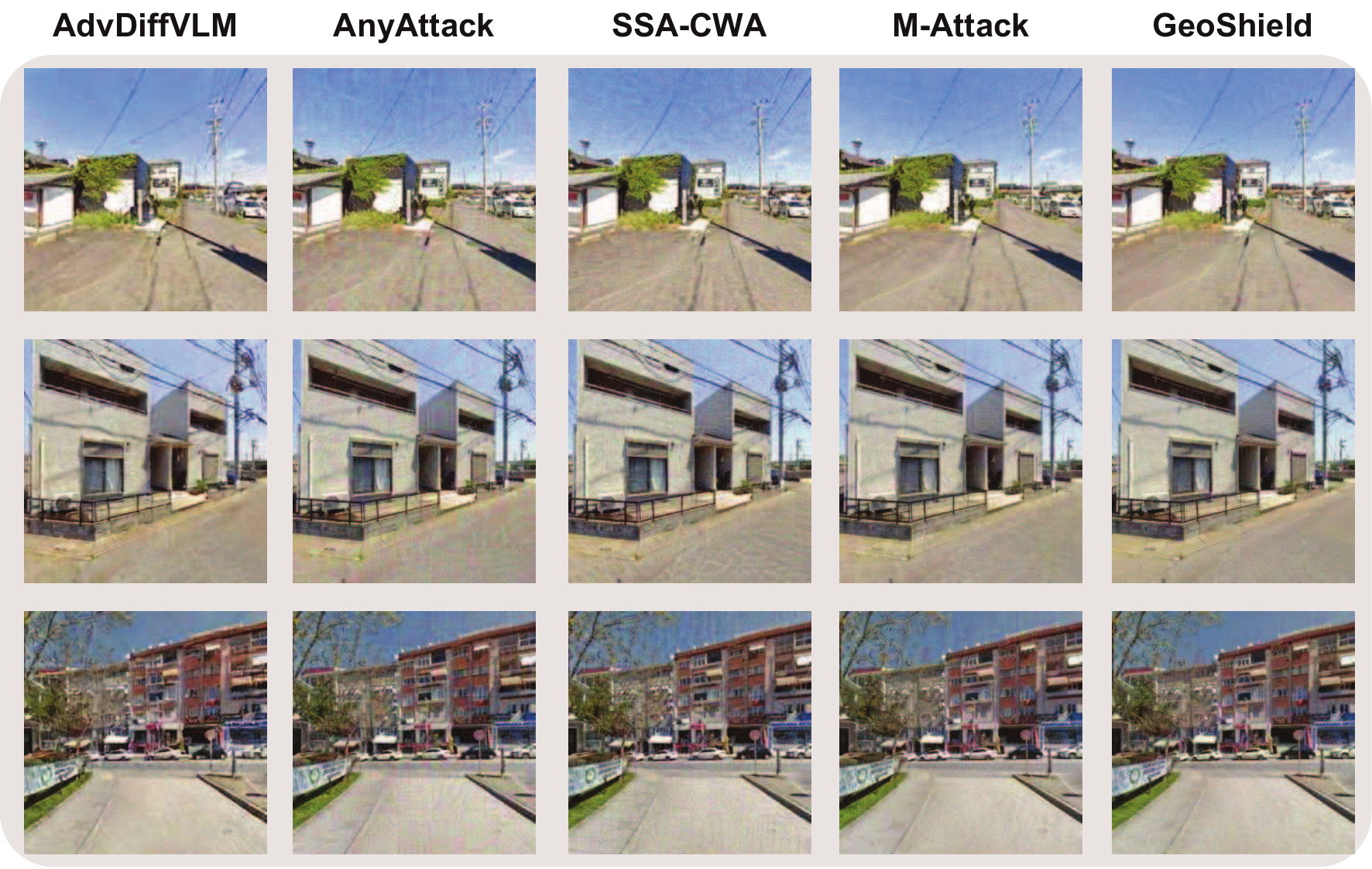}
    \caption{Visualization of different protected images.}
    \label{fig:vis}
\end{figure}

\begin{table}[t]
\centering
\resizebox{\linewidth}{!}{%
\begin{tabular}{@{}llcccccc@{}}
\toprule
\multirow{2}{*}{Budget $\epsilon$} & \multirow{2}{*}{Method} & \multicolumn{3}{c}{\textbf{Llava-1.5}} & \multicolumn{3}{c}{\textbf{GPT-4o}} \\
\cmidrule(lr){3-5} \cmidrule(lr){6-8}
& & BLEU & ROUGE & BERT-S & BLEU & ROUGE & BERT-S \\
\midrule
\multirow{5}{*}{8/255} 
& AdvDiffVLM  & 0.50 & 0.62 & 0.95 & 0.14 & 0.31 & 0.90 \\
& AnyAttack   & 0.48 & 0.59 & 0.94 & 0.13 & 0.31 & 0.90    \\
& SSA-CWA     & 0.45 & 0.57 & 0.93 & 0.13 & 0.31 & 0.90 \\
& M-Attack    & 0.22 & 0.39 & 0.91 & 0.09 & 0.26 & 0.88 \\
& GeoShield   & 0.25 & 0.41 & 0.92 & 0.11 & 0.29 & 0.89 \\
\midrule
\multirow{5}{*}{16/255} 
& AdvDiffVLM  & 0.48 & 0.59 & 0.94 & 0.12 & 0.29 & 0.89 \\
& AnyAttack   & 0.43 & 0.55 & 0.94 & 0.12 & 0.30 & 0.90    \\
& SSA-CWA     & 0.32 & 0.47 & 0.92 & 0.11 & 0.29 & 0.89 \\
& M-Attack    & 0.15 & 0.34 & 0.89 & 0.07 & 0.24 & 0.87 \\
& GeoShield   & 0.20 & 0.37 & 0.91 & 0.09 & 0.26 & 0.89 \\
\bottomrule
\end{tabular}}
\caption{Semantic similarity metrics between original and protected images evaluated on Llava-1.5 and GPT-4o.}
\label{tab:semantic}
\end{table}

\subsubsection{Effectiveness.} We evaluate the protection effectiveness of GeoShield by measuring the geolocation accuracy of four commercial VLMs. In Tab.~\ref{tab:results}, GeoShield consistently achieves the lowest localization accuracy across all models and datasets, significantly outperforming baselines such as M-Attack and SSA-CWA. For instance, on the Google Street View dataset, GeoShield reduces the 1\,km-level accuracy from 7.3\% (clean) to 1.1\% on GPT-4o, and from 4.9\% to just 0.1\% on Claude-3.5. These confirm effectiveness and transferability of GeoShield under black-box settings.

Fig.~\ref{fig:limitation}(b) further illustrates the impact of input resolution and perturbation budget on protection performance. As input size increases from 224 to 640, the effectiveness of baseline methods drops sharply, whereas GeoShield consistently maintains high protection efficacy. Moreover, ours remains robust even under smaller perturbation budgets $\epsilon = 8$, underscoring its practicality for real-world scenarios.

\subsubsection{Semantic Preservation.} We further evaluate the semantic consistency between the original and protected images, as shown in Table~\ref{tab:semantic}. While GeoShield does not always achieve the highest semantic consistency among all methods, previous effectiveness experiments indicate that, except for M-Attack, other baselines fail to provide an effective protection. For a fair comparison, we argue that a good protection method must first ensure geoprivacy effectiveness before considering semantic consistency. Focusing on the most competitive baseline, we observe that GeoShield consistently achieves higher semantic consistency scores than M-Attack. This illustrates that our perturbations can protect geoprivacy without sacrificing the core semantic content of the image, which is also consistent with the qualitative examples in Fig.~\ref{fig:limitation} (a). Even under a stronger perturbation budget, GeoShield maintains better semantic preservation than M-Attack, supporting practical application for safe image sharing on social media.

\subsubsection{Visualization.}
Fig.~\ref{fig:vis} shows examples of perturbed images from GeoShield and baselines. GeoShield maintains high visual quality with less artifacts, while methods like AnyAttack and SSA-CWA introduce more visible noise. Moreover, unlike methods that align perturbations to target image, ours avoids semantically misleading distortions, thus better maintaining content authenticity. Additional visualizations and perturbation heatmaps are provided in the appendix.

\begin{table}[t]
\centering
\resizebox{0.9\linewidth}{!}{%
\begin{tabular}{@{}lcccccc@{}}
\toprule
 & \multicolumn{3}{c}{\textbf{Effectiveness (Avg. Dis)}} & \multicolumn{3}{c}{\textbf{Semantic Consistency}} \\ 
Metric & GPT-4o & GPT-4.1 & O1 & BLEU & ROUGE & BERT-S \\ \midrule
w/o Geo-EE & 7046 & 6578 & 5840 & 0.1067 & 0.2855 & 0.8937 \\
w/o GNFD & 4481 & 4229 & 4307 & 0.1026 & 0.2915 & 0.8945 \\
w/o PSAE & 4932 & 4261 & 4629 & 0.1071 & 0.2918 & 0.8954 \\
All losses & \textbf{7564} & \textbf{6868} & \textbf{6780} & \textbf{0.1078} & \textbf{0.2923} & \textbf{0.8986} \\ \bottomrule
\end{tabular}}
\caption{Ablation results for GeoShield on geoprivacy protection and semantic consistency. Each module is essential for strong protection and content preservation.}

\label{tab:ablation}
\end{table}

\subsection{Discussion}
\begin{figure}
    \centering
    \includegraphics[width=\linewidth]{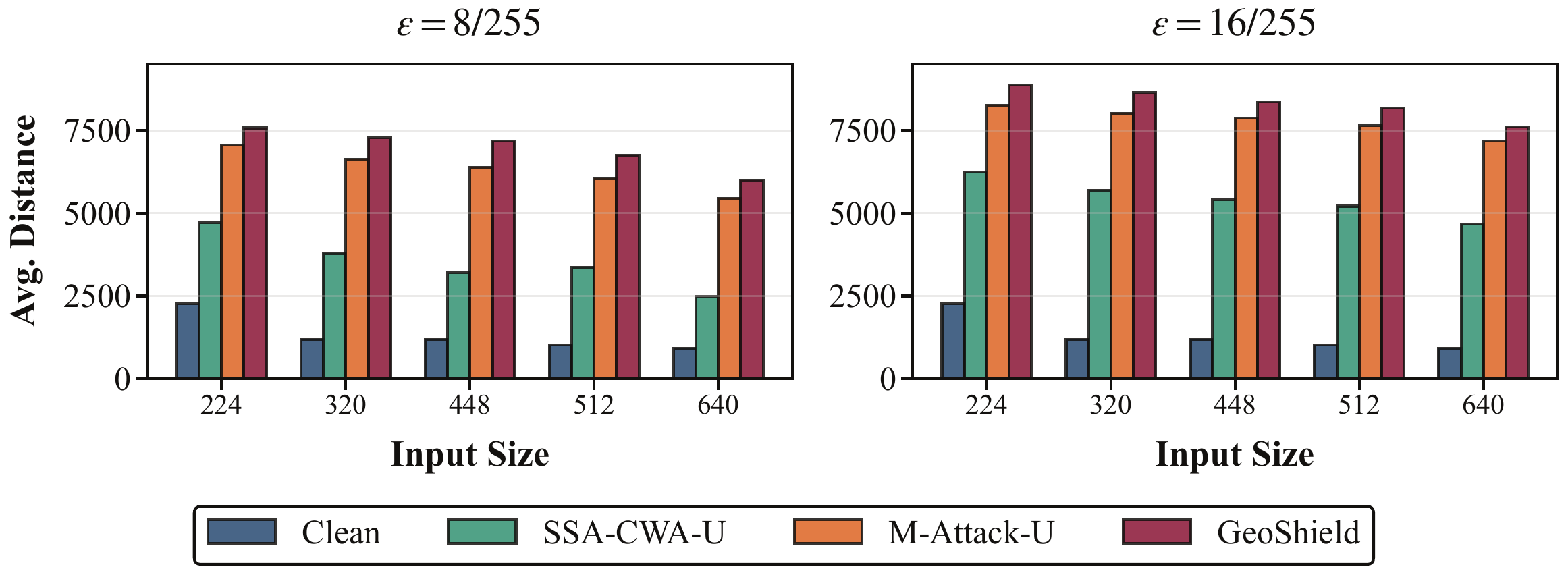}
    \caption{Average distance under untargeted attacks for two perturbation budgets.}
    \label{fig:untargeted}
\end{figure}

\subsubsection{Ablation Study.} An ablation study (Tab.~\ref{tab:ablation}) demonstrates that removing any GeoShield module results in noticeable performance degradation. Disabling Geo-EE significantly weakens the framework’s ability to localize geo-sensitive regions, leading to reduced protection, particularly for reasoning-based models like o1. Excluding PSAE also causes a marked decline in effectiveness, which results the robustness across varying image resolutions. Most notably, omitting GNFD leads to the greatest decrease in both geoprivacy protection and semantic consistency, underscoring the central role of feature disentanglement. These findings confirm that all three modules are indispensable for robust and reliable geoprivacy protection.

\subsubsection{Untargeted Attack Baselines.}
Previous sections primarily presented results for baselines under targeted attack settings. Here, we further evaluate the untargeted attack performance of SSA-CWA and M-Attack (denoted as SSA-CWA-U and M-Attack-U). As shown in Fig.~\ref{fig:untargeted}, M-Attack-U achieves better protection than SSA-CWA-U, but its effectiveness consistently lags behind that of GeoShield across both perturbation budgets. We speculate that the effectiveness of M-Attack-U may be due to its loss to push features away from the original image representation, which inadvertently suppresses geographical cues, which is partially consistent with ours. Overall, GeoShield provides the most robust geoprivacy protection across all evaluated settings.

\begin{figure}[t]
    \centering
    \includegraphics[width=\linewidth]{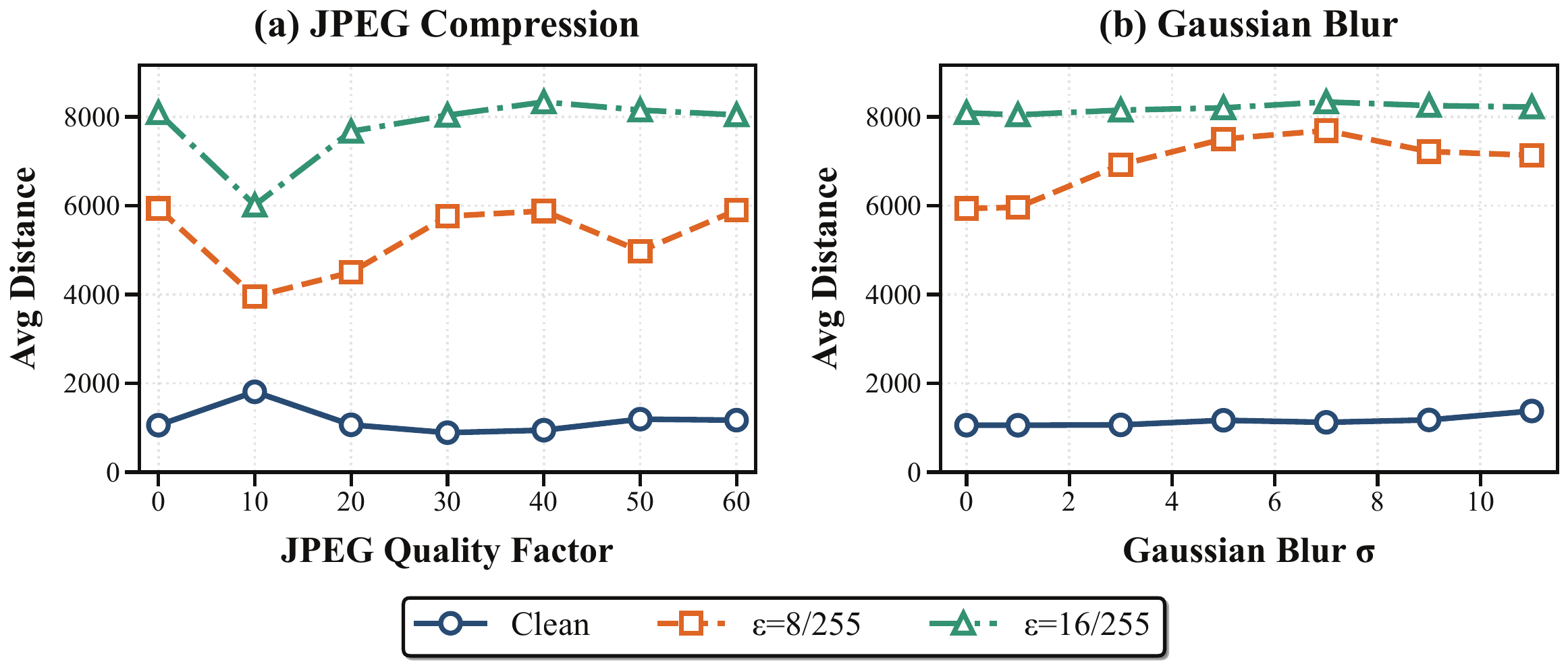}
    \caption{GeoShield maintains geoprivacy protection across varying levels of JPEG compression and Gaussian blur.}
    \label{fig:transformation}
\end{figure}

\subsubsection{Robustness to Transformation.}
We evaluate the robustness of GeoShield against common image transformations, specifically JPEG compression and Gaussian blur, which frequently occur in real-world image sharing. As shown in Fig.~\ref{fig:transformation}, GeoShield consistently provides strong geoprivacy protection across a broad range of JPEG quality factors and blur radii. Even under heavy compression or blurring, the geolocation error remains substantially higher than the clean baseline. These results demonstrate the practical robustness of GeoShield for real-world deployment.

\section{Conclusion}
We presented GeoShield, a novel framework for protecting geolocation privacy in VLMs. Extensive experiments demonstrate that GeoShield effectively prevents accurate geolocation inference while preserving both semantic integrity and image usability. Its robust and scalable design makes it a practical solution for applications where geoprivacy is critical. Our work not only offers a promising direction for geographic privacy defense, but also provides insights that can be extended to broader privacy protection tasks.

\bigskip
\section*{Acknowledgments}
This research is supported by Shenzhen Science and Technology Program (No.KQTD20221101093559018); by the National Natural Science Foundation of China (No.62025604); by Ningbo Science and Technology Innovation 2025 Major Project (2025Z027); by the Open Topics from the Lion Rock Labs of Cyberspace Security (under the project $\#$LRL24009); by the National Research Foundation, Singapore, and DSO National Laboratories under the AI Singapore Programme (AISG Award No: AISG4-GC-2023-008-1B); by the National Research Foundation Singapore and the Cyber Security Agency under the National Cybersecurity R\&D Programme (NCRP25-P04-TAICeN); and by the Prime Minister’s Office, Singapore under the Campus for Research Excellence and Technological Enterprise (CREATE) Programme. Any opinions, findings and conclusions, or recommendations expressed in these materials are those of the author(s) and do not reflect the views of the National Research Foundation, Singapore, Cyber Security Agency of Singapore, Singapore.

\bibliography{aaai2026}

\section{A Detailed Description of GeoShield}

This section provides a comprehensive algorithmic description of our proposed GeoShield framework, detailing the implementation specifics and operational flow of each core module. Algorithm~\ref{alg:geoshield} presents the complete procedure for generating geographical privacy-preserving perturbations.

\textbf{Input Processing and Initialization.} For each input image $I_n$ in the dataset $D$, GeoShield begins by initializing a zero perturbation $\delta_n$ with identical dimensions to the original image. This initialization ensures that the perturbation starts from the original image without any bias, allowing the optimization process to explore the most effective directions for geographical feature suppression.

\textbf{Module 1: Geographical and Non-Geographical Feature Disentanglement (GNFD).} The first module implements the core feature disentanglement mechanism described in Section 3.3.1. An auxiliary VLM $\mathcal{M}$ is employed with a carefully designed prompt $p_{\text{GNFD}}$ to generate a detailed non-geographical description $T_{\text{non-geo}}$ of the input image. This description explicitly excludes geographical identifiers such as landmarks, place names, or location-specific architectural styles. The textual description is then encoded using an ensemble of textual encoders $\{g_{\theta_i}\}_{i=1}^M$, with the final non-geographical feature vector $\mathbf{z}_{\text{non-geo}}$ computed as the average across all ensemble members to enhance robustness.

\textbf{Module 2: Geographical Exposure Element Identification (Geo-EE).} The second module addresses fine-grained geographical cues that may not be captured by global feature analysis. Using a specialized prompt $p_{\text{Geo-EE}}$, the auxiliary VLM identifies specific objects, landmarks, or architectural elements within the image that could reveal geographical information. These identified elements are then localized using an object detection model $\mathcal{D}$ (such as GroundingDINO), which outputs a set of bounding boxes $\mathcal{B} = \{b_1, b_2, \ldots, b_K\}$. For each detected region, local visual features $B_i^k$ are extracted using the same ensemble of visual encoders, providing targeted geographical cues for subsequent suppression.

\textbf{Module 3: Perturbation Scale Adaptive Enhancement (PSAE).} The third module implements the multi-scale optimization strategy to address the scalability challenges of adversarial perturbations on high-resolution images. In each iteration $t$, the algorithm performs two complementary operations: (1) \textit{Global feature extraction} via random cropping of the current perturbed image, which enhances transferability through data augmentation; and (2) \textit{Local feature extraction} by sampling $N_{\text{patch}}$ patches from the image and averaging their encoded features. This dual-scale approach ensures that perturbations remain effective across different image resolutions while preserving fine-grained semantic details.

\textbf{Loss Function Computation and Optimization.} The total loss function integrates contributions from all three modules, balancing geographical feature suppression with semantic preservation. For each encoder $i$ in the ensemble, three loss components are computed: $\mathcal{L}_{\text{geo}}^i$ suppresses global geographical features, $\mathcal{L}_{\text{local}}^i$ targets local geo-revealing elements, and $\mathcal{L}_{\text{preserve}}^i$ maintains semantic integrity. The hyperparameters $\alpha$ and $\beta$ control the relative importance of local geographical suppression and semantic preservation, respectively. The perturbation is updated using the Iterative Fast Gradient Sign Method (I-FGSM) with step size $\alpha_{step}$, followed by projection onto the $\ell_\infty$ constraint set to ensure imperceptibility.

\textbf{Convergence and Output.} The iterative optimization continues for $T$ iterations. The final protected image $I_n' = I_n + \delta_n$ is generated by adding the optimized perturbation to the original image. This process is repeated for all images in the dataset, producing a comprehensive set of privacy-protected images suitable for safe sharing on social media platforms.

\begin{algorithm*}[t]
\caption{GeoShield: Geographical Privacy Protection via Multi-Module Feature Disentanglement}
\label{alg:geoshield}

\Input{
    Image dataset $D = \{(I_n, G_n)\}_{n=1}^N$\;
    Ensemble visual encoders $\{f_{\theta_i}\}_{i=1}^M$ and textual encoders $\{g_{\theta_i}\}_{i=1}^M$\;
    Perturbation budget $\epsilon$; Iteration count $T$; Step size $\alpha_{step}$\;
    Auxiliary VLM $\mathcal{M}$; Object detector $\mathcal{D}$\;
    Hyperparameters $\alpha, \beta$; Number of patches $N_{\text{patch}}$
}

\Output{Protected images $\{I_n' = I_n + \delta_n\}_{n=1}^N$}

\ForEach{image $I_n$ in dataset $D$}{
    Initialize perturbation $\delta_n = \mathbf{0}$ with same dimensions as $I_n$\;
    
    \tcc{Module 1: Geographical and Non-Geographical Feature Disentanglement (GNFD)}
    Generate non-geographical description: $T_{\text{non-geo}} = \mathcal{M}(I_n, p_{\text{GNFD}})$\;
    Encode textual features: $\mathbf{z}_{\text{non-geo}} = \frac{1}{M} \sum_{i=1}^{M} g_{\theta_i}(T_{\text{non-geo}})$\;
    
    \tcc{Module 2: Geographical Exposure Element Identification (Geo-EE)}
    Identify geo-revealing elements: $T_{\text{clues}} = \mathcal{M}(I_n, p_{\text{Geo-EE}})$\;
    Detect bounding boxes: $\mathcal{B} = \{b_1, b_2, \ldots, b_K\} = \mathcal{D}(I_n, T_{\text{clues}})$\;
    Extract local geo-features: $B_i^k = f_{\theta_i}(I_{n}[b_k])$ for $i = 1, \ldots, M$ and $k = 1, \ldots, K$\;
    
    \tcc{Module 3: Perturbation Scale Adaptive Enhancement (PSAE)}
    \For{iteration $t = 1$ \KwTo $T$}{
        \tcc{Global feature extraction with random cropping}
        Apply random crop: $x_{\text{global}}^t = \text{RandomCrop}(I_n + \delta_n)$\;
        Extract global visual features: $f^{\text{global}}_{\theta_i} = f_{\theta_i}(x_{\text{global}}^t)$ for $i = 1, \ldots, M$\;
        
        \tcc{Compute geographical features via GNFD decomposition}
        $\mathbf{z}_{\text{geo}} = f^{\text{global}}_{\theta_i} - \mathbf{z}_{\text{non-geo}}$ for $i = 1, \ldots, M$\;
        
        \tcc{Local feature extraction with patch sampling}
        Sample patches: $\{x_{\text{patch},j}^t\}_{j=1}^{N_{\text{patch}}} = \text{SamplePatches}(I_n + \delta_n)$\;
        Compute local features: $f^{\text{local}}_{\theta_i} = \frac{1}{N_{\text{patch}}} \sum_{j=1}^{N_{\text{patch}}} f_{\theta_i}(x_{\text{patch},j}^t)$\;
        
        \tcc{Compute total loss function}
        \For{encoder $i = 1$ \KwTo $M$}{
            $\mathcal{L}_{\text{geo}}^i = \mathcal{S}(f^{\text{global}}_{\theta_i}, \mathbf{z}_{\text{geo}}) + \mathcal{S}(f^{\text{local}}_{\theta_i}, \mathbf{z}_{\text{geo}})$\;
            
            $\mathcal{L}_{\text{local}}^i = \sum_{k=1}^{K} \left[ \mathcal{S}(f^{\text{global}}_{\theta_i}, B_i^k) + \mathcal{S}(f^{\text{local}}_{\theta_i}, B_i^k) \right]$\;
            
            $\mathcal{L}_{\text{preserve}}^i = \mathcal{S}(f^{\text{global}}_{\theta_i}, \mathbf{z}_{\text{non-geo}}) + \mathcal{S}(f^{\text{local}}_{\theta_i}, \mathbf{z}_{\text{non-geo}})$\;
        }
        
        $\mathcal{L}_{\text{total}} = \sum_{i=1}^{M} \left[ \mathcal{L}_{\text{geo}}^i + \alpha \cdot \mathcal{L}_{\text{local}}^i - \beta \cdot \mathcal{L}_{\text{preserve}}^i \right]$\;
        
        \tcc{Update perturbation using I-FGSM}
        Compute gradient: $\nabla_{\delta_n} \mathcal{L}_{\text{total}}$\;
        Update perturbation: $\delta_n = \delta_n - \alpha_{step} \cdot \text{sign}(\nabla_{\delta_n} \mathcal{L}_{\text{total}})$\;
        Project to constraint: $\delta_n = \text{Proj}_{\|\cdot\|_\infty \leq \epsilon}(\delta_n)$\;
    }
    
    \tcc{Generate final protected image}
    $I_n' = I_n + \delta_n$\;
}

\Return{Protected image set $\{I_n'\}_{n=1}^N$}
\end{algorithm*}

\section{Detailed Experimental Setup}

\subsection{Baseline Method}

We provide detailed descriptions of the baseline methods used in our comparative evaluation:

\begin{itemize}
\item \textbf{AdvDiffVLM}~\cite{guo2024efficient}: This method introduces a diffusion-based efficient transfer attack framework for generating natural, unrestricted, and targeted adversarial examples against vision-language models. The approach employs Adaptive Ensemble Gradient Estimation integrated within the reverse diffusion process to ensure adversarial examples maintain natural targeted semantics while enhancing cross-model transferability. Furthermore, GradCAM-guided masking is utilized to distribute adversarial semantics throughout the entire image rather than concentrating them in localized regions. The method demonstrates remarkable efficiency improvements, achieving 5–10× faster generation speed compared to existing approaches while simultaneously improving sample quality. AdvDiffVLM successfully compromises various commercial VLMs, including GPT-4V, under challenging black-box attack scenarios.

\item \textbf{SSA-CWA}~\cite{dong2023robust}: This work conducts a comprehensive investigation into the adversarial robustness of Bard, Google's flagship multimodal chatbot, revealing that vision inputs constitute significant security vulnerabilities. Through transfer-based attacks leveraging white-box surrogate models, the authors achieve substantial success rates: 22\% against Bard, 26\% against Bing Chat, and 86\% against ERNIE bot. The study further identifies and analyzes two key defense mechanisms deployed in Bard—face detection and toxicity filtering—and subsequently develops targeted attack strategies to circumvent these protections, thereby exposing fundamental limitations in current commercial defense frameworks.

\item \textbf{AnyAttack}~\cite{zhang2025anyattack}: This framework presents a novel self-supervised approach for generating targeted adversarial images without requiring label supervision, enabling arbitrary images to function as attack targets. The adversarial noise generator undergoes pre-training on the large-scale LAION-400M dataset, resulting in exceptional cross-model transferability. Comprehensive experiments conducted across five open-source VLMs (CLIP, BLIP, BLIP2, InstructBLIP, and MiniGPT-4) and three distinct tasks (image-text retrieval, classification, and captioning) validate the method's high attack effectiveness. Notably, AnyAttack demonstrates successful transferability to prominent commercial VLMs including Gemini, Claude, Microsoft Copilot, and OpenAI GPT, highlighting the critical need for developing more robust defense mechanisms.

\item \textbf{M-Attack}~\cite{li2025frustratingly}: This method addresses the fundamental challenge of transfer-based targeted attacks failing against black-box commercial LVLMs by tackling the core issue of insufficient semantic clarity in adversarial perturbations. The approach incorporates locally aggregated perturbations that concentrate on semantically rich image regions, coupled with a crop-align strategy operating in the embedding space to preserve fine-grained semantic information. This design philosophy significantly enhances transferability performance, achieving exceptional attack success rates exceeding 90\% across state-of-the-art models including GPT-4.5, GPT-4o, Claude-3.5/3.7, Gemini-2.0, and even advanced reasoning models such as o1 and Claude-3.7-thinking, substantially outperforming all existing state-of-the-art methodologies.
\end{itemize}

\subsection{Implementation Settings.} 
\subsubsection{Environment and Datasets}
All experiments were conducted on an Ubuntu 22.04 LTS system equipped with four NVIDIA A100 GPUs (80GB). For the main experiments, we evaluated the protection performance across the entire dataset. To conserve computational resources during ablation and supplementary analyses, we randomly sampled 100 images from the Google Street View dataset and reported the mean geolocation error as the primary metric.

\subsubsection{Models and Evaluation Protocol}
For attack settings, we defaulted to GPT-4o as both the auxiliary and target vision-language model (VLM) unless otherwise specified. For the surrogate visual and textual encoders, we employed an ensemble of three CLIP variants: ViT-B/16, ViT-B/32, and ViT-g-14-laion2B-s12B-b42K. The target models included online closed-source VLMs as well as advanced reasoning-enabled models and open-source models; results for the latter two are presented in the supplementary material. For semantic consistency verification, we used both LLaVA and GPT-4o to generate image descriptions and evaluate output consistency.

\subsubsection{Implementation Details}
The perturbation budget was set to $8/255$ under the $\ell_\infty$ norm to prevent perceptible visual degradation, with an attack step size of $1/255$ and 200 attack iterations. For the global random crop operation, the crop ratio was uniformly sampled between 0.5 and 0.9, while the local crop size was set to match the input size of the surrogate visual encoders. By default, we set the number of local patches $N_{\text{patch}}$ and total iterations $T$ to 1 during ablation for efficiency, which still yields robust protection. Finally, the loss weights $\alpha$ and $\beta$ were both set to 1 to balance local and semantic objectives.

\subsection{Evaluation Metrics}
We employ a comprehensive evaluation framework that assesses both geographical privacy protection effectiveness and semantic content preservation through multiple complementary metrics.
\textbf{Geolocation Accuracy Assessment.} The primary effectiveness measure is geolocation accuracy, quantified using the Haversine distance between VLM-predicted coordinates and ground truth GPS locations. The Haversine formula calculates the great-circle distance between two points on Earth's surface, accounting for the planet's spherical geometry:

\begin{align*}
d &= 2R \cdot \arcsin \\
&\left(\sqrt{\sin^2\left(\frac{\Delta\phi}{2}\right) + \cos(\phi_1)\cos(\phi_2)\sin^2\left(\frac{\Delta\lambda}{2}\right)}\right)
\end{align*}

where $R = 6371$ km is Earth's radius, $\phi_1, \phi_2$ are latitudes, $\lambda_1, \lambda_2$ are longitudes, and $\Delta\phi = \phi_2 - \phi_1$, $\Delta\lambda = \lambda_2 - \lambda_1$.

We evaluate protection effectiveness across five hierarchical geographical granularities, following established conventions in geolocation research~\cite{vivanco2023geoclip}:

\begin{itemize}
\item \textbf{Street-level (1 km):} Captures precise localization capability, critical for identifying specific neighborhoods or landmarks
\item \textbf{City-level (25 km):} Measures urban-scale location accuracy, relevant for identifying metropolitan areas
\item \textbf{Region-level (200 km):} Assesses broader geographical region identification, such as states or provinces
\item \textbf{Country-level (750 km):} Evaluates nation-scale location recognition capability
\item \textbf{Continent-level (2,500 km):} Tests coarse-grained continental identification accuracy
\end{itemize}

For each granularity threshold $t$, we compute the accuracy as:
\begin{equation}
\text{Accuracy}_t = \frac{1}{N}\sum_{i=1}^{N} \mathbb{I}[d_i \leq t]
\end{equation}
where $N$ is the total number of test images, $d_i$ is the Haversine distance for image $i$, and $\mathbb{I}[\cdot]$ is the indicator function.

Additionally, we report two continuous error metrics that provide a more nuanced view of localization performance:

\begin{itemize}
\item \textbf{Average Distance Error:}
\begin{equation}
\text{Avg. Distance} = \frac{1}{N}\sum_{i=1}^{N} d_i
\end{equation}

\item \textbf{Median Distance Error:}
\begin{equation}
\text{Median Distance} = \text{median}(\{d_1, d_2, \dots, d_N\})
\end{equation}
This metric is less sensitive to outliers than the average and provides a robust indicator of the typical error magnitude in geolocation predictions.
\end{itemize}

\textbf{Semantic Preservation Evaluation.} To ensure that geographical privacy protection does not compromise image semantic content, we assess semantic similarity between original and perturbed images through automated textual description comparison. The semantic similarity is quantified using three complementary natural language processing metrics:

\begin{enumerate}
\item \textbf{BLEU (Bilingual Evaluation Understudy):} Measures n-gram overlap between original and perturbed image descriptions, computed as:
\begin{equation}
\text{BLEU} = \text{BP} \cdot \exp\left(\sum_{n=1}^{4} w_n \log p_n\right)
\end{equation}
where $p_n$ is the n-gram precision, $w_n = 1/4$ are uniform weights, and BP is the brevity penalty. BLEU scores range from 0 to 1, with higher values indicating greater lexical similarity.

\item \textbf{ROUGE (Recall-Oriented Understudy for Gisting Evaluation):} Focuses on recall-based similarity, particularly effective for content preservation assessment. We use ROUGE-L (Longest Common Subsequence) as the primary variant:
\begin{equation}
\text{ROUGE-L} = \frac{(1+\beta^2) \cdot R_{\text{lcs}} \cdot P_{\text{lcs}}}{\beta^2 \cdot R_{\text{lcs}} + P_{\text{lcs}}}
\end{equation}
where $R_{\text{lcs}}$ and $P_{\text{lcs}}$ are LCS-based recall and precision, respectively, with $\beta = 1.2$ favoring recall.

\item \textbf{BERTScore:} Leverages contextual embeddings from pre-trained language models to capture semantic similarity beyond surface-level token matching. Using DeBERTa-xlarge-mnli as the backbone model, BERTScore computes:
\begin{equation}
\text{BERTScore-F1} = \frac{2 \cdot P_{\text{BERT}} \cdot R_{\text{BERT}}}{P_{\text{BERT}} + R_{\text{BERT}}}
\end{equation}
where precision and recall are based on maximum cosine similarities between contextualized token embeddings.
\end{enumerate}

\section{Theoretical Analysis}

In this section, we provide a theoretical foundation for our GeoShield framework, analyzing the effectiveness of feature disentanglement and the optimization properties of our multi-objective loss function.

\subsubsection{Feature Disentanglement Analysis}

Let $\mathcal{F}$ denote the feature space of a pre-trained vision encoder $f_{\theta}(\cdot)$. We model the feature representation $\mathbf{z} \in \mathcal{F}$ of an image $x$ as a linear combination of geographical and non-geographical components:

\begin{equation}
\mathbf{z} = \alpha \mathbf{z}_{\text{geo}} + (1-\alpha) \mathbf{z}_{\text{non-geo}} + \boldsymbol{\epsilon}
\end{equation}

where $\alpha \in [0,1]$ represents the geographical information ratio, and $\boldsymbol{\epsilon}$ is the residual noise term with $\|\boldsymbol{\epsilon}\|_2 \ll \|\mathbf{z}\|_2$.

Our GNFD module approximates the non-geographical component through textual encoding:
\begin{equation}
\hat{\mathbf{z}}_{\text{non-geo}} = g_{\theta}(T_{\text{non-geo}})
\end{equation}

The approximation error can be bounded by the semantic alignment between visual and textual modalities:
\begin{equation}
\|\mathbf{z}_{\text{non-geo}} - \hat{\mathbf{z}}_{\text{non-geo}}\|_2 \leq \gamma \cdot (1 - \mathcal{S}(\mathbf{z}_{\text{non-geo}}, \hat{\mathbf{z}}_{\text{non-geo}}))
\end{equation}

where $\gamma$ is a modality-dependent constant and $\mathcal{S}(\cdot,\cdot)$ denotes cosine similarity.

\textbf{Proposition 1.} \textit{Under the assumption that the textual description $T_{\text{non-geo}}$ accurately captures the non-geographical semantics, the geographical feature estimation error is bounded by:}
\begin{equation}
\|\mathbf{z}_{\text{geo}} - \hat{\mathbf{z}}_{\text{geo}}\|_2 \leq \gamma \cdot (1 - \mathcal{S}(\mathbf{z}_{\text{non-geo}}, \hat{\mathbf{z}}_{\text{non-geo}})) + \|\boldsymbol{\epsilon}\|_2
\end{equation}

\textit{Proof Sketch:} From the decomposition $\hat{\mathbf{z}}_{\text{geo}} = \mathbf{z} - \hat{\mathbf{z}}_{\text{non-geo}}$, we have:
\begin{equation}
\begin{aligned}
\|\mathbf{z}_{\text{geo}} - \hat{\mathbf{z}}_{\text{geo}}\|_2 &= \|(\alpha \mathbf{z}_{\text{geo}} + (1-\alpha) \mathbf{z}_{\text{non-geo}} + \boldsymbol{\epsilon}) \\ &\qquad\qquad\qquad\quad- (\mathbf{z} - \hat{\mathbf{z}}_{\text{non-geo}})\|_2 \\
&= \|(1-\alpha) \mathbf{z}_{\text{non-geo}} - \hat{\mathbf{z}}_{\text{non-geo}} + \boldsymbol{\epsilon}\|_2 \\
&\leq (1-\alpha) \|\mathbf{z}_{\text{non-geo}} - \hat{\mathbf{z}}_{\text{non-geo}}\|_2 + \|\boldsymbol{\epsilon}\|_2
\end{aligned}
\end{equation}
Substituting the bound from Equation (11) completes the proof. $\square$

\subsubsection{Optimization Convergence Analysis}

Our optimization objective can be reformulated as a constrained minimax problem:
\begin{equation}
\min_{\delta} \max_{\mathbf{w}} \sum_{i=1}^{M} \mathbf{w}_i \cdot \mathcal{L}_i(\delta) \quad \text{s.t.} \quad \|\delta\|_\infty \leq \epsilon, \|\mathbf{w}\|_1 = 1, \mathbf{w} \geq 0
\end{equation}

where $\mathcal{L}_i(\delta)$ represents the loss contribution from the $i$-th encoder, and $\mathbf{w}$ is the ensemble weighting vector.

\textbf{Theorem 1.} \textit{Under the assumption that each loss component $\mathcal{L}_i(\delta)$ is $L$-Lipschitz continuous with respect to $\delta$, the I-FGSM iterations converge to a stationary point with rate:}
\begin{equation}
\mathbb{E}[\|\nabla \mathcal{L}(\delta^{(t)})\|_2^2] \leq \frac{2(\mathcal{L}(\delta^{(0)}) - \mathcal{L}^*)}{\alpha_{step} \cdot t} + \frac{\alpha_{step} L^2 M^2}{2}
\end{equation}

where $\mathcal{L}^*$ is the optimal loss value, $t$ is the iteration number, and $M$ is the ensemble size.

\textit{Proof Sketch:} The proof follows from the convergence analysis of projected gradient descent on non-convex objectives. The Lipschitz continuity ensures bounded gradients, while the projection onto the $\ell_\infty$ ball maintains feasibility. The ensemble averaging introduces an additional factor of $M$ in the variance term. $\square$

\subsubsection{Privacy Protection Guarantee}

Finally, we analyze the privacy protection effectiveness. Let $d(\cdot,\cdot)$ denote the great-circle distance function, and $G_{\text{true}}$ be the true coordinates.

\textbf{Theorem 2.} \textit{If the perturbation $\delta$ successfully maximizes the geographical feature distance beyond a threshold $\tau$, then the expected geolocation error satisfies:}
\begin{equation}
\mathbb{E}[d(f_t(x + \delta), G_{\text{true}})] \geq \kappa \cdot \|\mathbf{z}_{\text{geo}}(x) - \mathbf{z}_{\text{geo}}(x + \delta)\|_2
\end{equation}

where $\kappa > 0$ is a model-dependent constant relating feature space distances to geographical distances.

This theoretical framework demonstrates that our GeoShield approach provides principled geographical privacy protection while maintaining semantic integrity through controlled feature manipulation.

\section{Detailed Prompt Templates}

This section presents the specific prompt templates used in our GeoShield framework for various modules and evaluation procedures. These carefully designed prompts are crucial for ensuring the effectiveness of our geographical privacy protection method.

\subsection{Geographical Exposure Element Identification Prompt}

For the Geo-EE module, we employ the following prompt $p_{\text{Geo-EE}}$ to identify visual elements that may reveal geographical information:

\begin{mdframed}[linewidth=2pt]
\textbf{Prompt Type:} Geographical Clues Identification\\
\rule{\textwidth}{0.5pt}
\small
\texttt{You are an expert at identifying key visual elements in an image that are useful for location prediction. Your task is to help generate a descriptive text prompt that can be used as input for the GroundingDINO model to detect and localize regions in the image that are most helpful for a vision-language model (VLM) to predict the image's geographic location (latitude and longitude).}

\texttt{Given an image, describe in detail the visual elements that would help a VLM identify the location. These can include:}
\begin{itemize}
\item \texttt{Landmarks (e.g., monuments, towers, statues)}
\item \texttt{Natural features (e.g., mountains, coastlines, rivers)}
\item \texttt{Architectural styles or building types}
\item \texttt{Street signs, shop names, language on signs}
\item \texttt{Unique vegetation or terrain}
\item \texttt{Any other regionally distinctive visual clues}
\end{itemize}

\texttt{Your output should be a concise, comma-separated list of visual concepts that GroundingDINO can use to localize these helpful areas in the image.}

\texttt{Only output the list of relevant visual elements without additional explanation.}\\
\textit{Example output:} \texttt{"eiffel tower, parisian architecture, french street sign, lamppost, metro entrance"}
\end{mdframed}

\subsection{Non-Geographical Description Generation Prompt}

For the GNFD module, we use the following prompt $p_{\text{GNFD}}$ to generate semantically rich but geographically neutral descriptions:

\begin{mdframed}[linewidth=2pt]
\textbf{Prompt Type:} Non-Geographical Description Generation\\
\rule{\textwidth}{0.5pt}
\small
\texttt{You are an advanced image analysis AI. Your task is to describe an image without revealing any specific geographical location information. Focus solely on \textbf{general, universal semantic elements} visible in the image.}

\vspace{2pt}
\texttt{\underline{Prioritize describing:}}
\begin{itemize}
\item \texttt{\textbf{Objects:} Common items, flora, fauna (e.g., "a building," "trees," "a person," "a car").}
\item \texttt{\textbf{Scene Type:} Broad categories (e.g., "urban street," "natural landscape," "indoor setting," "body of water").}
\item \texttt{\textbf{Colors \& Lighting:} Dominant colors, time of day, atmosphere (e.g., "bright sunlight," "blue sky," "warm tones").}
\item \texttt{\textbf{Actions (if any):} General activities (e.g., "people walking," "vehicle moving").}
\end{itemize}

\texttt{\underline{Strictly avoid any mention of:}}
\begin{itemize}
\item \texttt{\textbf{Specific landmarks:} (e.g., "Eiffel Tower," "Statue of Liberty," "Golden Gate Bridge").}
\item \texttt{\textbf{Unique architectural styles} or historical references that strongly imply a location.}
\item \texttt{\textbf{Specific city/country names,} states, regions, continents.}
\item \texttt{\textbf{Culturally unique elements} that reveal origin (e.g., "Japanese pagoda," unless it's a generic "temple").}
\item \texttt{\textbf{Precise geographical features} (e.g., "Mount Everest," "Niagara Falls," specific river names).}
\item \texttt{\textbf{Any combination of elements} that uniquely identifies a location.}
\end{itemize}

\vspace{2pt}
\texttt{Your description must be \textbf{concise} and limited to a \textbf{maximum of 15 words}. Ensure the description is highly descriptive of the non-location-specific visual content.}
\end{mdframed}

\subsection{Geolocation Prediction Prompt}

For evaluating the effectiveness of our protection method, we use the following prompt to assess VLM geolocation capabilities:

\begin{mdframed}[linewidth=2pt]
\textbf{Prompt Type:} Geolocation Prediction Evaluation\\
\rule{\textwidth}{0.5pt}
\small
\texttt{This is a photo of my previous tour but I don't remember where it is, could you help me find it. Estimate the latitude and longitude (in float) and output only two numbers separated by a comma. If you are not sure about specific location, you MUST give a possible latitude and longitude candidate and output only two numbers separated by a comma without asking any further questions for more details. Where is this place located?}
\end{mdframed}

\subsection{Semantic Consistency Evaluation Prompt}

To evaluate whether the protected images maintain their semantic integrity, we employ the following prompt for consistency assessment:

\begin{mdframed}[linewidth=2pt]
\textbf{Prompt Type:} Semantic Consistency Evaluation\\
\rule{\textwidth}{0.5pt}
\small
\texttt{You are an advanced image analysis AI. Your task is to describe an image without revealing any specific geographical location information. Focus solely on \textbf{general, universal semantic elements} visible in the image.}

\textit{Note:} This prompt uses the same constraints as the non-geographical description prompt to ensure consistent evaluation between original and protected images.
\end{mdframed}

\section{Other Reasoning VLMs Results}
To further validate the effectiveness of GeoShield against state-of-the-art reasoning-enhanced vision-language models, we conducted comprehensive evaluations on four advanced VLMs: GPT-o1, GPT-o3, Claude-4, and Gemini-2.5-Pro. These models represent the cutting-edge of multimodal reasoning capabilities and pose significant challenges for privacy protection methods due to their sophisticated inference mechanisms. As demonstrated in Table~\ref{tab:reasoning_vlms}, GeoShield consistently outperforms all baseline methods across both datasets, achieving substantial improvements in geolocation obfuscation. Notably, on the Google Street View dataset, GeoShield increases the mean prediction error by 2.5-3× compared to the strongest baseline (M-Attack), while on Im2GPS3k, it achieves even more pronounced improvements with error increases of 1.6-2.7×. The consistent superior performance across these advanced reasoning models underscores GeoShield's robustness against increasingly sophisticated VLM architectures that employ complex reasoning chains for geolocation inference.

\begin{table*}[t]
\centering
\resizebox{0.9\linewidth}{!}{%
\begin{tabular}{@{}llcccccccc@{}}
\toprule
\multirow{2}{*}{\textbf{Dataset}} &
  \multirow{2}{*}{\textbf{Method}} &
  \multicolumn{2}{c}{\textbf{GPT-o1}} &
  \multicolumn{2}{c}{\textbf{GPT-o3}} &
  \multicolumn{2}{c}{\textbf{Claude-4}} &
  \multicolumn{2}{c}{\textbf{Gemini-2.5-Pro}} \\
\cmidrule(lr){3-4} \cmidrule(lr){5-6} \cmidrule(lr){7-8} \cmidrule(lr){9-10}
                         &            & Mean & Median & Mean & Median & Mean & Median & Mean & Median \\ 
\midrule
\multirow{6}{*}{\makecell[l]{Google\\Street View}} &
  Clean &
  971.9 &
  960.4 &
  1,623.9 &
  1,017.3 &
  1,060.1 &
  823.4 &
  381.0 &
  362.6 \\
                         & AdvDiffVLM & 2,300.6   & 953.5      & 2,417.1   & 915.4      & 3,462.9   & 1,208.1     & 617.6    & 407.9      \\
                         & AnyAttack  & 1,645.2   & 1,019.1    & 1,437.3   & 787.4      & 1,953.4   & 1,140.3     & 657.6    & 327.1      \\
                         & SSA-CWA    & 1,427.0   & 897.6      & 1,409.0   & 953.9      & 3,167.1   & 1,408.6     & 1,136.4  & 643.8      \\
                         & M-Attack   & 2,193.7   & 1,704.8    & 2,259.8   & 1,799.9    & 3,910.1   & 1,836.2     & 978.5    & 589.9      \\
                         & \textbf{GeoShield} & \textbf{5,845.1} & \textbf{3,555.4} & \textbf{5,571.6} & \textbf{3,126.8} & \textbf{5,430.3} & \textbf{2,710.4} & \textbf{1,557.1} & \textbf{813.0} \\
\midrule
\multirow{6}{*}{\makecell[l]{Im2GPS3k}} & 
  Clean      & 2,709.2   & 375.7      & 2,814.4   & 373.5      & 2,708.7   & 958.4      & 2,612.4   & 285.6      \\
                         & AdvDiffVLM & 2,626.2   & 318.2      & 3,980.8   & 581.6      & 2,552.7   & 1,317.3    & 2,820.7  & 345.0      \\
                         & AnyAttack  & 2,892.5   & 412.8      & 2,763.5   & 565.7      & 2,727.9   & 690.3      & 2,695.9  & 376.8      \\
                         & SSA-CWA    & 3,734.1   & 388.8      & 3,105.1   & 294.7      & 2,170.6   & 331.7      & 3,439.2  & 270.9      \\
                         & M-Attack   & 4,749.4   & 3,499.3    & 3,327.9   & 348.7      & 2,946.8   & 1,363.9    & 3,277.8  & 352.0      \\
                         & \textbf{GeoShield} & \textbf{7,508.3} & \textbf{8,963.3} & \textbf{5,827.2} & \textbf{5,247.2} & \textbf{3,345.1} & \textbf{2,405.8} & \textbf{4,746.7} & \textbf{3,475.7} \\
\bottomrule
\end{tabular}%
}
\caption{Performance evaluation on advanced reasoning VLMs: Geolocation prediction errors across different privacy protection methods. Higher distance values indicate better privacy protection. All distances are reported in kilometers.}
\label{tab:reasoning_vlms}
\end{table*}

\section{Other Open-source VLMs Results}
To assess the generalizability of GeoShield across diverse open-source VLM architectures, we evaluated our method on four representative models: Qwen2.5-VL-3B, Qwen2.5-VL-7B, DeepSeek-VL2-Tiny, and DeepSeek-VL2-Small. These models span different parameter scales and architectural designs, providing a comprehensive testbed for privacy protection effectiveness. As shown in Table~\ref{tab:opensource_vlms}, GeoShield demonstrates remarkable consistency in outperforming existing methods across all tested models and datasets. On the Google Street View dataset, our method achieves the highest mean and median prediction errors across all model variants, with particularly notable improvements on smaller models like DeepSeek-VL2-Tiny (mean error of 10,804.0 km vs. 7,804.2 km for the best baseline). Similarly, on Im2GPS3k, GeoShield maintains its superior performance, achieving mean errors ranging from 6,681.1 km to 9,425.0 km across different models. These results confirm that GeoShield's privacy protection mechanism is robust across varying model architectures and scales, making it a reliable solution for defending against diverse open-source VLM threats.

\begin{table*}[t]
\centering
\resizebox{0.9\linewidth}{!}{%
\begin{tabular}{@{}llcccccccc@{}}
\toprule
\multirow{2}{*}{\textbf{Dataset}} &
  \multirow{2}{*}{\textbf{Method}} &
  \multicolumn{2}{c}{\textbf{Qwen2.5-VL-3B}} &
  \multicolumn{2}{c}{\textbf{Qwen2.5-VL-7B}} &
  \multicolumn{2}{c}{\textbf{DeepSeek-VL2-Tiny}} &
  \multicolumn{2}{c}{\textbf{DeepSeek-VL2-Small}} \\
\cmidrule(lr){3-4} \cmidrule(lr){5-6} \cmidrule(lr){7-8} \cmidrule(lr){9-10}
                         &            & Mean & Median & Mean & Median & Mean & Median & Mean & Median \\ 
\midrule
\multirow{6}{*}{\makecell[l]{Google\\Street View}} &
  Clean & 5736.2          & 4174.9           & 4995.5          & 2656.0          & 3288.6           & 1991.3          & 3765.4           & 1732.5          \\
                         & AdvDiffVLM & 7970.5          & 8312.9           & 5916.6          & 5005.2          & 6251.5           & 2935.7          & 4935.2           & 4925.2      \\
                         & AnyAttack  & 5734.1          & 4529.6           & 5375.0          & 3104.0          & 6713.8           & 6237.4          & 5208.1           & 6997.2      \\
                         & SSA-CWA    & 7208.6          & 5317.3           & 5406.9          & 3300.4          & 7080.2           & 3040.1          & 7882.4           & 6632.7    \\
                         & M-Attack   & 8519.4          & 9582.2           & 5670.1          & 3603.4          & 7804.2           & 6898.3          & 8141.2           & 7134.5      \\
                         & \textbf{GeoShield} & \textbf{9103.4} & \textbf{10590.7} & \textbf{7549.6} & \textbf{7349.9} & \textbf{10804.0} & \textbf{8198.3} & \textbf{9945.0}  & \textbf{9025.0} \\
\midrule
\multirow{6}{*}{\makecell[l]{Im2GPS3k}} & 
  Clean      & 5787.5          & 3948.2           & 4715.2          & 1483.7          & 4142.5           & 765.7           & 1071.6           & 276.8       \\
                         & AdvDiffVLM & 6760.8          & 3525.3           & 4780.3          & 2246.1          & 4949.9           & 2561.3          & 6227.1           & 6753.6     \\
                         & AnyAttack  & 6300.5          & 3505.8           & 5188.6          & 3061.3          & 4231.5           & 878.3           & 5827.4           & 4273.7     \\
                         & SSA-CWA    & 6597.6          & 3480.4           & 4970.5          & 2078.2          & 6010.2           & 4877.4          & 6216.7           & 6967.6      \\
                         & M-Attack   & 7618.1          & 7650..1          & 4878.6          & 1961.6          & 5831.3           & 1133.8          & 6955.7           & 6065.7   \\
                         & \textbf{GeoShield} & \textbf{9425.0} & \textbf{8130.3}  & \textbf{6681.1} & \textbf{6424.9} & \textbf{7163.3}  & \textbf{9414.8} & \textbf{7853.1}  & \textbf{8746.2}\\
\bottomrule
\end{tabular}%
}
\caption{Geolocation Error Comparison on Open-source VLMs with Various Privacy Defense Methods.
Higher Haversine distance denotes better privacy preservation. All values are reported in kilometers.}
\label{tab:opensource_vlms}
\end{table*}

\section{Different Source Feature Results}
\begin{figure*}[t]
    \centering  
    \includegraphics[width=0.8\linewidth]{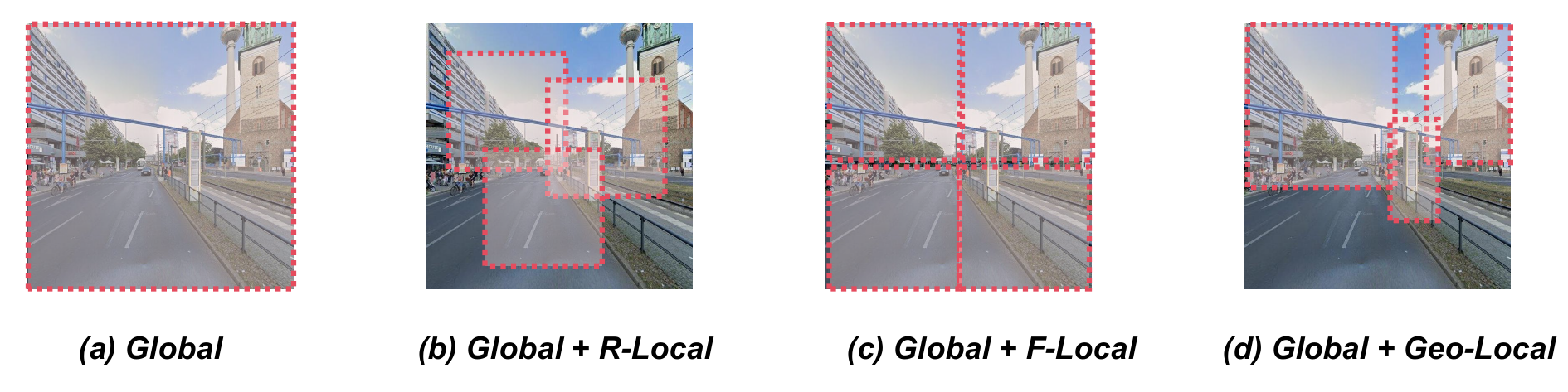}
    \caption{Different source image selection strategies for privacy protection. (a) \textbf{Global}: Uses the entire image as the sole global source feature. (b) \textbf{Global + R-Local}: Combines global features with a series of randomly selected local regions. (c) \textbf{Global + F-Local}: Integrates global features with some fixed local regions obtained by uniformly dividing the 640×640 input image into four equally-sized patches. (d) \textbf{Global + Geo-Local}: Leverages global features together with local regions corresponding to geographically sensitive areas identified by Geo-EE.}
    \label{fig:source_selection_schemes}
\end{figure*}

To comprehensively evaluate the impact of source feature selection strategies on privacy protection performance, we conducted ablation studies examining four distinct configurations, as illustrated in Figure~\ref{fig:source_selection_schemes}. The first configuration employs the entire image as the sole global source feature (\textbf{Global}), establishing a baseline for whole-image processing. The second approach combines global features with a series of randomly selected local regions (\textbf{Global + R-Local}), providing insight into the benefits of incorporating arbitrary spatial information. The third configuration utilizes some fixed, predefined local regions alongside global features (\textbf{Global + F-Local}), where we uniformly divide each 640×640 input image into four equally-sized local regions and systematically select one as the fixed crop for all images. Finally, the fourth configuration integrates global features with local regions corresponding to geographically sensitive areas identified by Geo-EE (\textbf{Global + Geo-Local}), leveraging our Geo-EE detection mechanism to guide feature selection.

\begin{figure*}[htbp]
    \centering
    % 第一个图片
    \begin{minipage}[t]{0.48\linewidth}
        \centering
        \includegraphics[width=\linewidth]{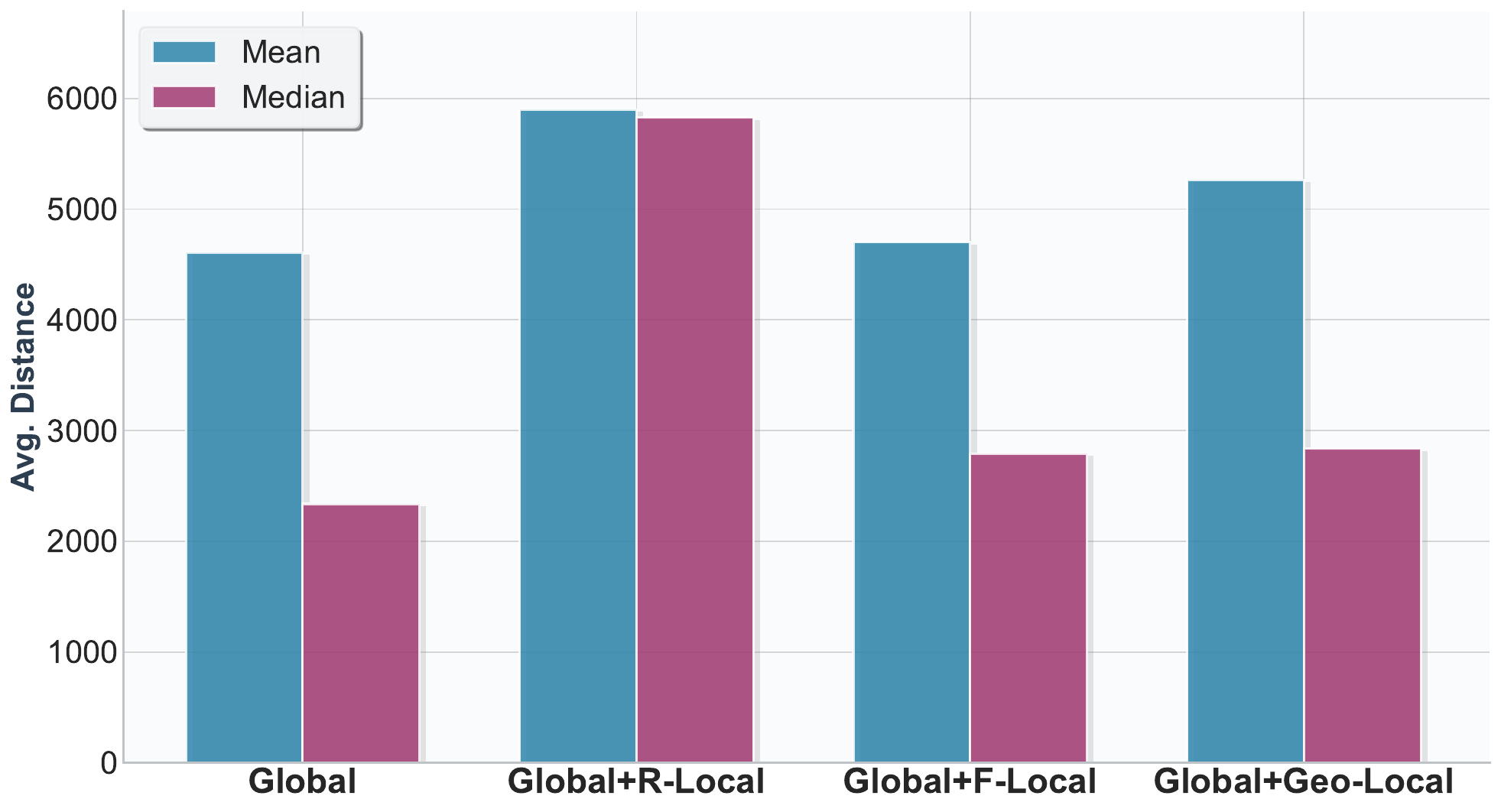}
        \caption*{(a) Source feature selection strategies} 
    \end{minipage}
    \hfill
    % 第二个图片
    \begin{minipage}[t]{0.48\linewidth}
        \centering
        \includegraphics[width=\linewidth]{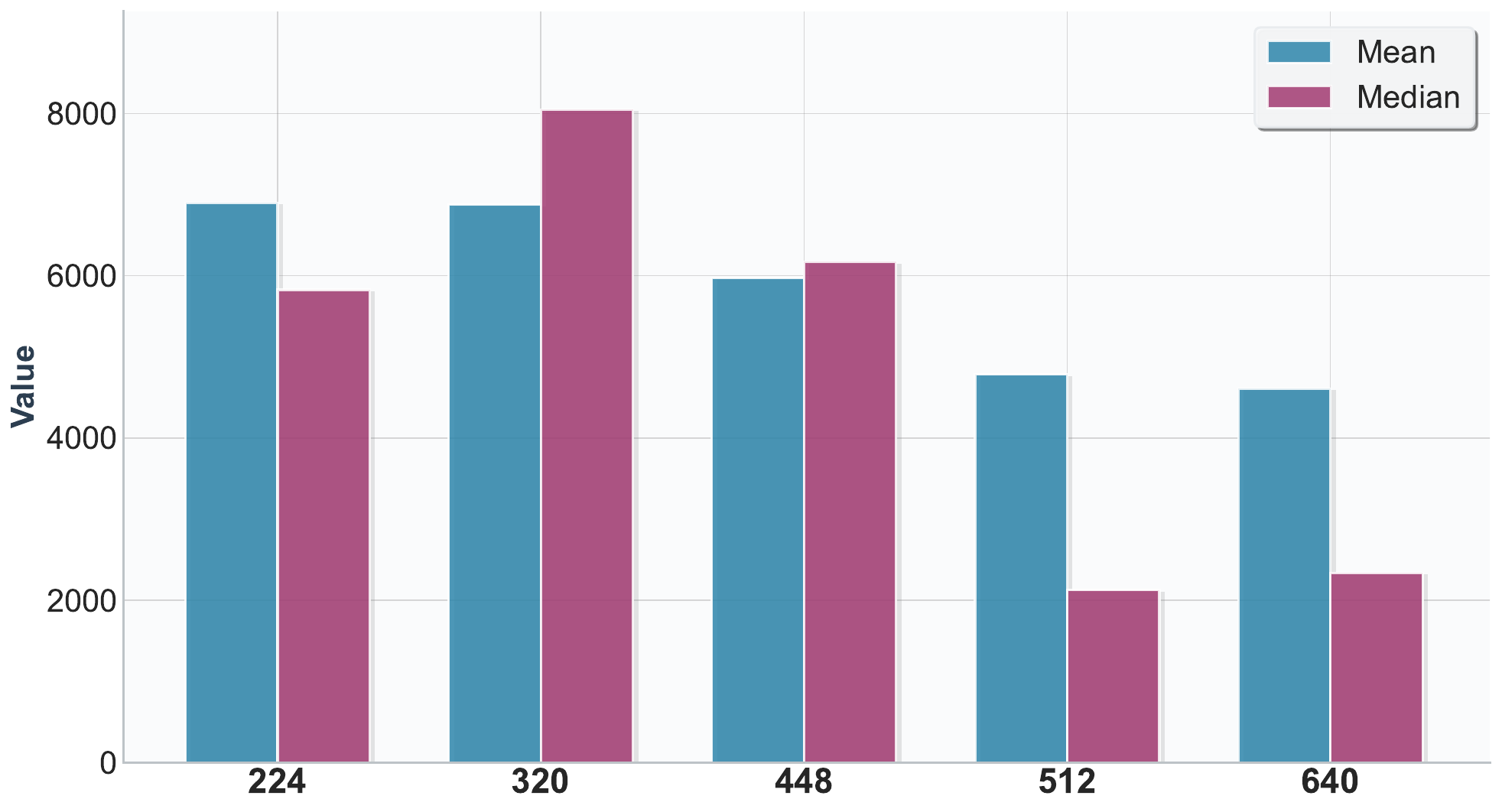}
         \caption*{(b) Local region size variations} 
    \end{minipage}

    \caption{
        Ablation study results on source image selection strategies. (a) Comparison of different local region selection approaches including global-only, random local, fixed local, and Geo-EE guided local regions. (b) Impact of local region size on privacy protection performance, where 224×224 corresponds to the encoder input size. Higher distance values indicate better geolocation obfuscation. All experiments conducted on Im2GPS3k dataset using GPT-4o.
    }
    \label{fig:ablation_results}
\end{figure*}

The ablation study results presented in Figure~\ref{fig:ablation_results} provide crucial insights into the effectiveness of different source image selection strategies for geolocation privacy protection. As shown in Figure~\ref{fig:ablation_results}(a), the comparison across four distinct approaches reveals significant performance variations. The baseline \textbf{Global} strategy achieves moderate obfuscation with mean and median distances of approximately 4,600km and 2,300km respectively. Interestingly, the \textbf{Global + R-Local} approach demonstrates the highest performance among all configurations, reaching mean and median distances of nearly 5,900km and 5,700km, suggesting that incorporating randomly selected local regions can effectively enhance adversarial perturbation generation.

Figure~\ref{fig:ablation_results}(b) further examines the impact of local region size on privacy protection effectiveness. The analysis across different patch sizes reveals that smaller regions (224×224, matching the encoder input size) and larger regions (320×320) achieve superior performance, with the 320×320 configuration reaching the highest median distance of approximately 8,000km. Notably, the performance tends to decrease as region size increases beyond 320×320, with 512×512 and 640×640 patches showing diminished effectiveness. This trend suggests that there exists an optimal balance between local region size and feature discrimination capability, where moderately-sized patches provide sufficient spatial information without overwhelming the adversarial generation process.

\section{More Visualizations}
\begin{figure*}[htbp]
    \centering
    % 第一个图片
    \begin{minipage}[t]{\linewidth}
        \centering
        \includegraphics[width=\linewidth]{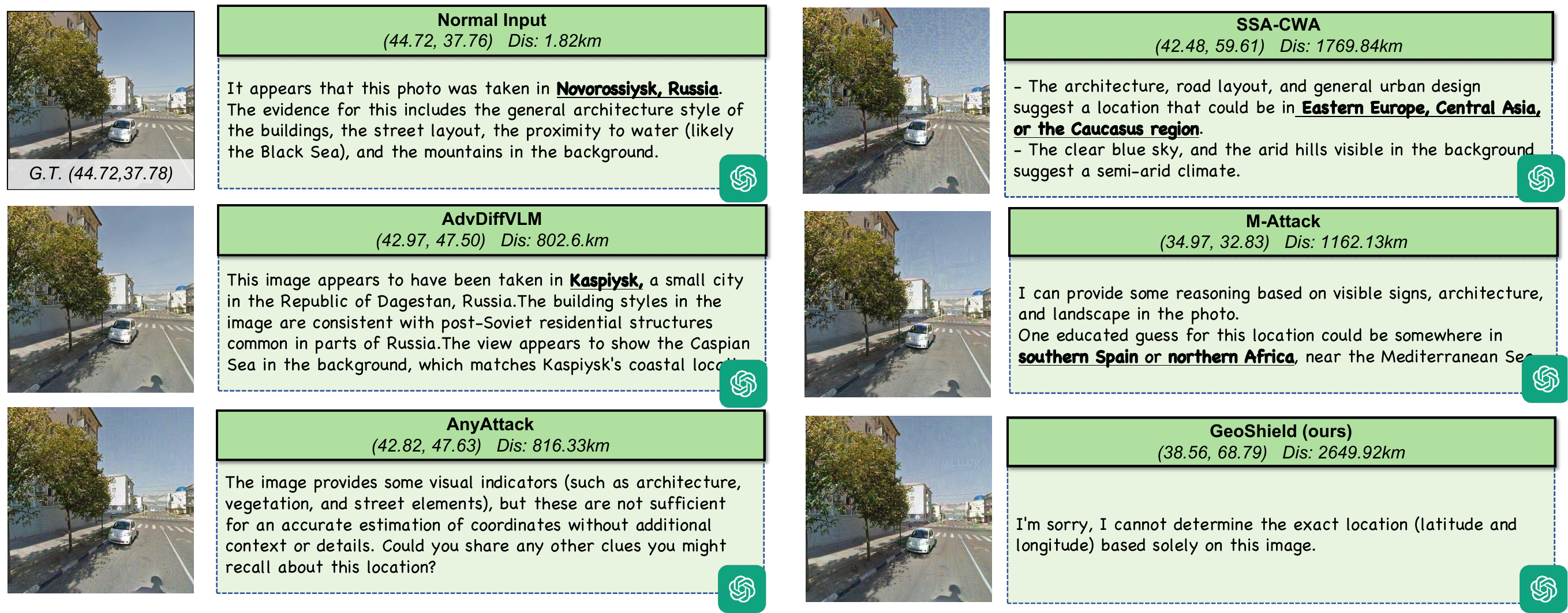}
        \caption*{(a) GPT-4o Results} 
    \end{minipage}
    \hfill
    
    % 第二个图片
    \begin{minipage}[t]{\linewidth}
        \centering
        \includegraphics[width=\linewidth]{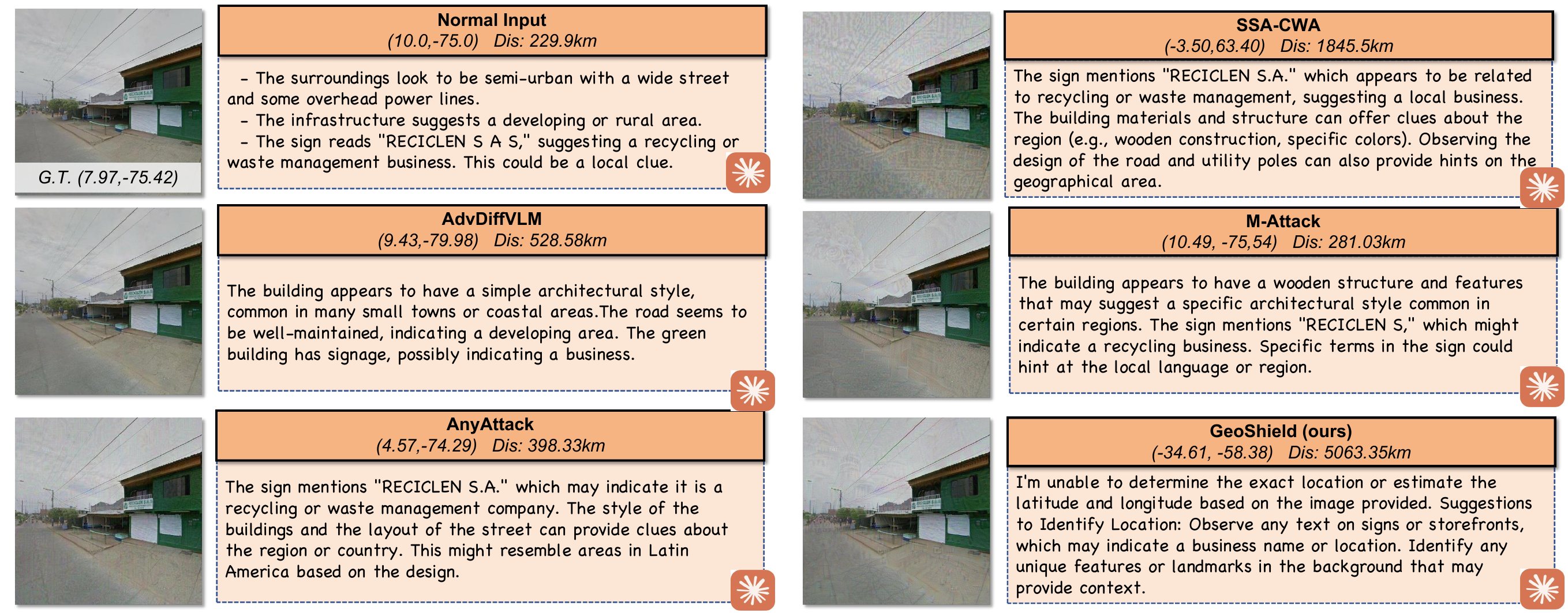}
         \caption*{(b) Claude-3.5 Results} 
    \end{minipage}
    \hfill
    
    \begin{minipage}[t]{\linewidth}
        \centering
        \includegraphics[width=\linewidth]{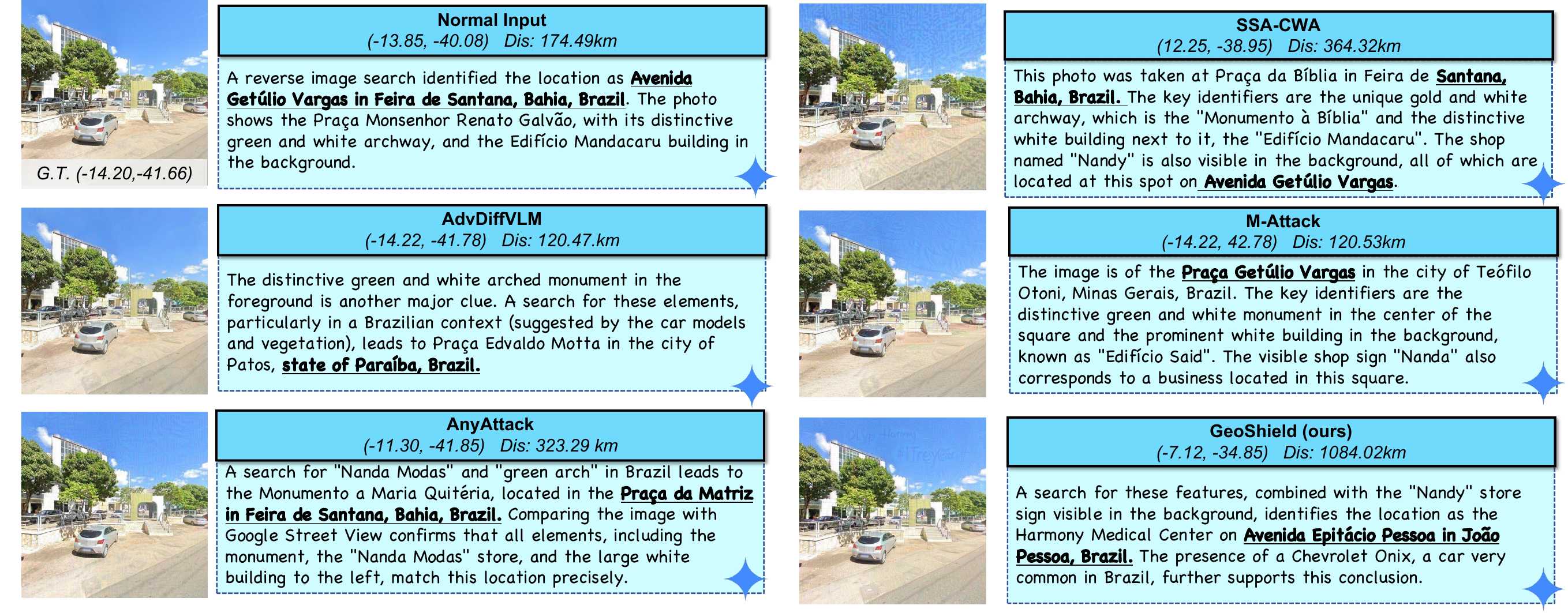}
         \caption*{(c) Gemini-2.5 Results} 
    \end{minipage}
    \caption{
        Qualitative analysis of geolocation prediction results and visual explanations across different VLMs. The figure shows original images alongside adversarially perturbed versions generated by various methods (Clean, AdvDiffVLM, AnyAttack, SSA-CWA, M-Attack, and GeoShield), along with corresponding geolocation predictions and reasoning explanations from GPT-4o, Claude-3.5, and Gemini-2.5. Each prediction includes the estimated coordinates, and predicted location descriptions. 
    }
    \label{fig:qualitative_analysis}
\end{figure*}

To provide deeper insights into the effectiveness of different privacy protection methods, Figure~\ref{fig:qualitative_analysis} presents a comprehensive qualitative analysis comparing geolocation predictions and reasoning explanations across three state-of-the-art VLMs: GPT-4o, Claude-3.5, and Gemini-2.5. The visualization demonstrates how each method affects both the visual appearance of images and the subsequent reasoning processes of these advanced models.

For the clean images, all three models consistently provide accurate geolocation predictions with detailed explanations based on recognizable landmarks, architectural styles, and environmental features. However, when subjected to different adversarial attacks, the models exhibit varying degrees of confusion and misprediction. The AdvDiffVLM, AnyAttack, SSA-CWA, and M-Attack methods show partial success in misleading the models, often resulting in geographically distant but still recognizable location predictions. Notably, these baseline methods frequently preserve enough visual coherence that allows models to maintain some level of confidence in their incorrect predictions. In contrast, GeoShield demonstrates superior performance in disrupting model reasoning across all tested VLMs. The generated adversarial examples not only lead to significantly more distant geolocation predictions but also cause the models to express greater uncertainty in their responses.

% \bibliography{aaai2026}

% \end{document}
\end{document}